\documentclass[11pt]{article}

\ifdefined\NCEREVIEW
  \usepackage[review]{acl}
\else
  \usepackage[preprint]{acl}
\fi

\usepackage{times}
\usepackage{latexsym}
\usepackage[T1]{fontenc}
\usepackage[utf8]{inputenc}
\usepackage{microtype}
\usepackage{inconsolata}
\usepackage{graphicx}
\graphicspath{{Figures/}}

\usepackage{xurl}
\usepackage{amsmath}
\usepackage{amssymb}
\usepackage{booktabs}
\usepackage{array}
\usepackage{colortbl}
\usepackage{tikz}
\usetikzlibrary{calc,backgrounds,arrows.meta}
\usepackage{fontawesome5}
\usepackage{tcolorbox}

\definecolor{nceAddrBg}{HTML}{DCE7F7}
\definecolor{nceAddrFg}{HTML}{1C4A86}
\definecolor{nceEnvBg}{HTML}{3A4352}
\definecolor{nceDropBg}{HTML}{F7D3D1}
\definecolor{nceDropFg}{HTML}{A1282A}
\definecolor{nceRewBg}{HTML}{FAE1B0}
\definecolor{nceRewFg}{HTML}{7A4A00}
\definecolor{nceAppBg}{HTML}{D6EDDA}
\definecolor{nceAppFg}{HTML}{2C6A3B}

\makeatletter
\newcommand{\nceRel}[1]{\fontsize{\strip@pt\dimexpr#1\dimexpr\f@size pt\relax\relax}{\f@baselineskip}\selectfont}
\makeatother
\tikzset{nce chip/.style={rounded corners=1.6pt, inner xsep=1.7pt, inner ysep=0pt,
  text height=1.6ex, text depth=0.42ex, font=\nceRel{0.9}\ttfamily}}
\DeclareRobustCommand{\nceChip}[3]{\mbox{\tikz[baseline=(c.base)]\node[nce chip,fill=#1,text=#2](c){#3};}}
\newcommand{\nceIcon}[1]{{\nceRel{0.75}#1}\,}
\DeclareRobustCommand{\opdrop}{\nceChip{nceDropBg}{nceDropFg}{\nceIcon{\faIcon[regular]{trash-alt}}drop}}
\DeclareRobustCommand{\oprewrite}{\nceChip{nceRewBg}{nceRewFg}{\nceIcon{\faPen}rewrite}}
\DeclareRobustCommand{\opappend}{\nceChip{nceAppBg}{nceAppFg}{\nceIcon{\faIcon[regular]{plus-square}}append}}
\DeclareRobustCommand{\tokctx}{\nceChip{nceEnvBg}{white}{<ctx>}}
\DeclareRobustCommand{\tokctxend}{\nceChip{nceEnvBg}{white}{</ctx>}}
\DeclareRobustCommand{\tokaddr}[2]{\nceChip{nceAddrBg}{nceAddrFg}{<sid\_hi\_#1>}\hspace{0.15em}\nceChip{nceAddrBg}{nceAddrFg}{<sid\_lo\_#2>}}

\definecolor{nceTakeBg}{HTML}{EDF3FB}
\definecolor{nceTakeFg}{HTML}{24508A}
\definecolor{nceLinkFg}{HTML}{2563EB}
\definecolor{nceLinkTx}{HTML}{1D4ED8}
\DeclareRobustCommand{\nceLink}[3]{%
  \href{#3}{{\color{nceLinkFg}#1}{\hspace{0.42em}\fontfamily{phv}\selectfont\fontsize{8.6}{9}\selectfont
   \raisebox{0.135em}{\textcolor{nceLinkTx}{#2}}}}}
\newcommand{\nceLinkSep}{\hspace{1.05em}{\color{black!22}\rule[-0.18em]{0.45pt}{1.05em}}\hspace{1.05em}}

\newtcolorbox{takeaway}{colback=nceTakeBg, colframe=nceTakeFg, boxrule=0pt, leftrule=2pt,
  sharp corners, left=5pt, right=5pt, top=3pt, bottom=3pt, before skip=7pt, after skip=7pt,
  fontupper=\small, before upper={{\color{nceTakeFg}\faLightbulb[regular]~\textbf{Takeaway.}}~}}

\title{Editing as Language: Teaching LLM Agents a Grammar for\\Native Context Editing}

\ifdefined\NCEREVIEW\else
  \hypersetup{pdfkeywords={context editing, long-horizon agents, LLM agents,
    agentic context management, special tokens, distillation}}
\fi

\newcommand{\blfootnote}[1]{\begingroup
  \renewcommand{\thefootnote}{}%
  \footnotetext{\raggedright #1}%
  \addtocounter{footnote}{-1}\endgroup}

\author{
\ifdefined\NCEREVIEW
Anonymous authors
\else
\vspace{3mm}
{\textbf{\href{https://openreview.net/profile?id=~Yunlong_Deng1}{Yunlong Deng}}$^{1,3,*}$\quad
 \textbf{\href{https://openreview.net/profile?id=~Yan_Li44}{Yan Li}}$^{2,*}$\quad
 \href{https://openreview.net/profile?id=~Xing_Hu7}{Xing Hu}$^{3}$\quad
 \href{https://openreview.net/profile?id=~Yida_Zhu1}{Yida Zhu}$^{3}$}\\[0.2em]
{\href{https://openreview.net/profile?id=~Sean_Du1}{Sean Du}$^{2}$\quad
 \textbf{\href{https://openreview.net/profile?id=~Weiyang_Liu1}{Weiyang Liu}}$^{1,\dag}$\quad
 \textbf{\href{https://openreview.net/profile?id=~Guangyi_Chen1}{Guangyi Chen}}$^{4,5,\dag}$}\\[0.7em]
{\normalsize\normalfont
 $^{1}$The Chinese University of Hong Kong\quad
 $^{2}$Nanyang Technological University\quad
 $^{3}$Meituan LongCat Team\\[0.15em]
 $^{4}$Mohamed bin Zayed University of Artificial Intelligence\quad
 $^{5}$Carnegie Mellon University}
\fi
}

\begin{document}

\ifdefined\NCEREVIEW\else
\makeatletter
\renewcommand{\@maketitle}{%
  \vbox{%
    \hsize\textwidth \linewidth\hsize
    \vskip -0.6in
    \noindent
    \begin{minipage}[b]{0.2\textwidth}%
      \href{https://longcat.chat}{\includegraphics[width=\linewidth]{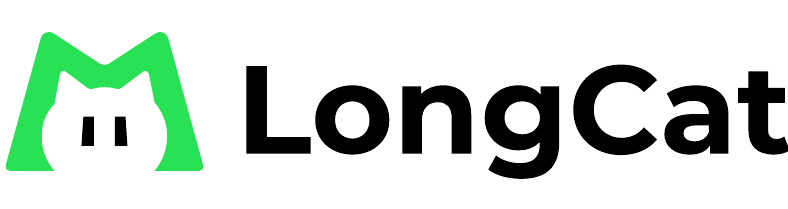}}%
    \end{minipage}%
    \par\vspace{-11pt}
    \noindent\rule{\linewidth}{1pt}%
    \vspace{15pt}
    \centering
    {\LARGE\bfseries\@title\par}%
    \vspace{0.75em}
    {\large\@author\par}%
    \vspace{0.85em}%
    \nceLink{\faGithub}{Code}{https://github.com/dengyl20/native-context-editing}%
    \nceLinkSep
    \nceLink{\faGlobe}{Project page}{https://dengyl20.github.io/native-context-editing/}%
    \par
  }%
  \vskip 0.25in
}
\makeatother
\fi

\maketitle
\ifdefined\NCEREVIEW\else
\blfootnote{$^{*}$Equal contribution.\quad $^{\dag}$Corresponding authors. Correspondence to
\href{mailto:1155281107@link.cuhk.edu.hk}{\mbox{1155281107@link.cuhk.edu.hk}} (Y.~Deng),
\href{mailto:wyliu@cse.cuhk.edu.hk}{\mbox{wyliu@cse.cuhk.edu.hk}} (W.~Liu), and
\href{mailto:guangyichen1994@gmail.com}{\mbox{guangyichen1994@gmail.com}} (G.~Chen).
Y.~Deng's work was done during an internship at Meituan.}
\fi

\begin{abstract}

A synopsis in novels folds a story into a few lines; a function in code gathers many steps with a single execution. Context editing is native to language itself. In this paper, we propose the Native Context Editing (NCE) framework, which gives language model agents a grammar for natively editing their own context. By encoding context-editing operations in a grammar of dedicated special tokens, we teach the agents to invoke edits through the language itself. We train agents on iteratively constructed trajectory segments, with each teacher edit conditioned on the evolving context. This establishes context editing as an atomic and inherent capability of LLM agents, enabling them to dynamically modify their context as they generate responses over long-horizon reasoning and interaction. Compared with context-editing methods that learn to generate tool-use code within specific external harness environments, our approach integrates context editing as an internal mechanism of the model’s native generation process, enabling better generalization across diverse real-world workflows.
On four long-horizon agentic benchmarks (Terminal-Bench 2.0, SWE-bench Verified, SWE-bench Multilingual, and SWE-bench Pro), NCE improves the average task success rate over the base model by 5.6 percentage points for LongCat-Flash-Lite-Sparse and by 4.5 for Qwen3.6-35B-A3B, while reducing average token usage by 36\% and 15\%, respectively.
\end{abstract}

\section{Introduction}
\label{sec:intro}

\begin{figure*}[t]
  \centering
  \includegraphics[width=1\textwidth]{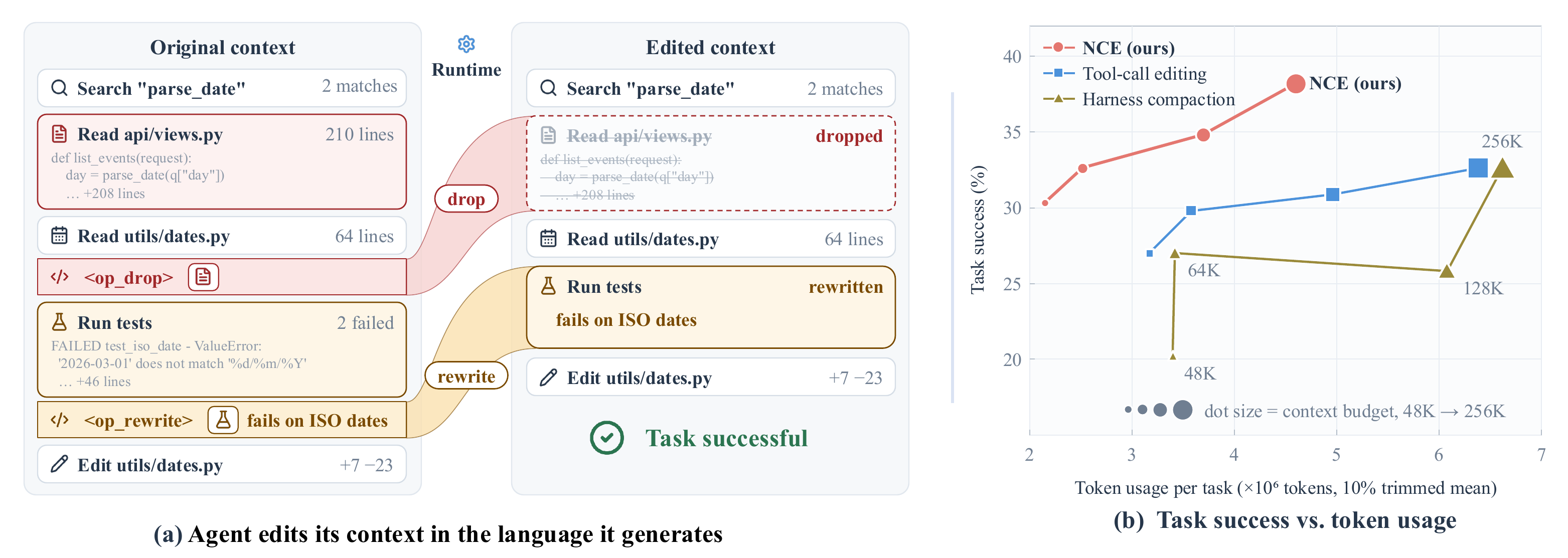}
  \caption{\textbf{Overview of NCE.} (a)~The agent writes edits inline with its output, each naming a target block by its icon: \opdrop{} removes a file it no longer needs, \oprewrite{} condenses a failing test log to one line; the runtime applies them before the agent continues. (b)~Success vs.\ tokens per task for LongCat-Flash-Lite-Sparse (69B-A3B) on Terminal-Bench 2.0 (Table~\ref{tab:tb_budgets}). NCE succeeds most with the fewest tokens at every budget.}
  \label{fig:main}
\end{figure*}

\providecommand{\pend}[1][XX]{\textcolor{red}{[#1]}}
Large language model (LLM) agents now tackle tasks that require many rounds of reasoning, tool use, and interaction with environments~\citep{sweagent, openhands, openclaw-rl}.
In these long-horizon tasks, the interactions accumulate code, tool outputs, execution feedback, and intermediate decisions into increasingly long histories~\citep{complexitytrap}. 
Long-horizon agents therefore need to maintain useful working context within a finite window: retain evidence for upcoming decisions, shorten completed investigations, and remove information that is no longer needed.

Existing approaches manage context through either predefined rules or external tools.
Even the most widely deployed coding agents rely on rules. Claude Code and Codex summarize the history once it crosses a fixed token threshold~\citep{claudecode,codex}, and simpler policies mask older observations~\citep{complexitytrap}. These triggers fire on context size alone and cannot tell which information the task still needs.
Tool-based methods expose context management through external tool calls, letting the agent decide what to edit. Sculptor provides tools to fragment, summarize, hide, and restore the history~\citep{sculptor}, and Context-Folding folds a finished sub-trajectory into a summary~\citep{contextfolding}.
Because the harness defines and executes these tools, the learned editing policy is bound to harness-specific interfaces and generalizes poorly across different environments.

Both families define context editing outside the model. Rules fix the policy in the harness, and tools, even when a learned policy calls them, take their meaning from schemas that the harness declares in the prompt. We therefore seek to answer:
\begin{quote}
\itshape 
How can we turn context editing into a native capability of language models?
\end{quote}

We call context editing native when each edit takes its meaning from the model's own vocabulary, as the end-of-sequence token <eos> does. No prompt needs to explain <eos>, and it ends generation the same way in every harness. Our answer is to make editing part of the language the model speaks. Language already provides natural primitives for organizing information, \emph{e.g.}, a summary compresses a story into a few lines, a function name abstracts a sequence of operations, and an updated note supersedes outdated information. Guided by this intuition, we introduce Native Context Editing (NCE), a framework that equips LLM agents with an executable grammar to compress, abstract, update, and restructure their own context as reasoning unfolds (Figure~\ref{fig:main}a).

Specifically, NCE organizes the context into addressable blocks, and its grammar has three components: a \emph{vocabulary} of dedicated tokens, in which address tokens name blocks and operator tokens (\opdrop, \oprewrite, \opappend) remove, replace, or add them; \emph{composition rules} that define well-formed edits; and \emph{semantics} that bind each well-formed edit to exactly one transition of the context state, executed by a task-agnostic runtime. The meaning of every edit thus ships with the model and holds in every harness.
Recorded trajectories contain no edits, so we annotate them with two teachers. An offline teacher reads the full trajectory and assigns each block a release point, the turn after which the block may be edited. A block is released only when the teacher can cite the event that ended its use. An online teacher then uses the previous blocks and release annotations from the offline teacher to decide when and how to edit. The edits are thus safe to learn and executable by an agent that cannot see ahead.

We then train NCE through distillation over these editing trajectories, with supervision that follows the context as it is edited. Specifically, each edit is executed at the end of the turn that issues it, and every later editing decision and task action is constructed and supervised on the resulting context. Annotating all edits on the original history would condition later decisions on text that the agent has already deleted or rewritten, a mismatch that grows with every edit. 
By executing each edit before the next turn is built, NCE trains the agent on contexts shaped by its earlier edits. This is exactly the condition it faces at inference time, therefore bridging the train-test gap. During inference, the model generates each turn, the runtime executes the edits it contains, and generation resumes from the edited context, so the context can evolve throughout long-horizon reasoning.

In the experiments, we have implemented NCE on LongCat-Flash-Lite-Sparse (69B-A3B) and Qwen3.6-35B-A3B and evaluated both on four long-horizon agentic benchmarks, including Terminal-Bench 2.0~\citep{terminalbench}, SWE-bench Verified~\citep{swebench}, SWE-bench Multilingual~\citep{swemultilingual}, and SWE-bench Pro~\citep{swepro}. On LongCat at 256K, NCE improves success over the base model by 4.8 to 6.3 points and reduces token usage by 31\% to 44\% on every benchmark (Table~\ref{tab:main_results} and Figure~\ref{fig:main}b). A tool-call editing baseline trained on the same edits trails NCE by 5.2 to 8.0 points, so the gap comes from expressing the edits natively. The gains transfer to Qwen3.6, where NCE raises the average success rate over the base model by 4.5 points and cuts average token usage by 15\%.

We summarize our contributions below:
\begin{itemize}
\setlength{\itemsep}{2pt}\setlength{\parskip}{0pt}
\item We introduce Native Context Editing (NCE), a framework that teaches LLM agents a grammar to dynamically manage their own active context through natively generated edits.
\item We develop a distillation framework with evolving contexts, where an offline teacher uses release analysis in hindsight to decide which blocks to edit, and an online teacher that sees only the present makes the edits.
\item We evaluate NCE on four long-horizon agentic benchmarks, demonstrating its effectiveness in improving task performance while also reducing both context and token usage over the base model with harness compaction.
\end{itemize}

\section{Related Work}
\label{sec:related}

\paragraph{Long-horizon LLM agents.}
Research on long-horizon LLM agents has investigated how interaction with
external environments can support reasoning and decision-making. For example,
ReAct interleaves reasoning and action, while SWE-agent and OpenHands
provide interfaces for code editing and
execution~\citep{react,sweagent,openhands}. Recent long-horizon agents
learn to consolidate memory or fold their context as interactions
accumulate~\citep{mem1,agentfold}.
These approaches depend on retaining useful evidence across decisions, yet
longer histories do not ensure reliable evidence access~\citep{lostmiddle}.
NCE addresses this shared challenge by maintaining the context available to
ongoing reasoning and interaction.

\paragraph{Memory management and knowledge retrieval for LLM agents.}
Retrieval-augmented generation and agentic memory systems store information outside the
context window and retrieve it whenever later decisions need
it~\citep{rag,memgpt,amem}. Memory managers decide what and when to store, update, or
discard, often with database-like operations such as SQL or
ADD/UPDATE/DELETE~\citep{chatdb,memoryr1}, and some apply these decisions
to the active context through tool calls~\citep{memact,statelm,agemem}. Unlike most memory management methods, NCE edits the active textual context of an ongoing task without introducing an external knowledge storage or information retrieval mechanism. 
Our approach is a parallel contribution to external memory, focusing specifically on active-context management within a fixed context-window budget.

\paragraph{Context management for LLM agents.}
Rule-based methods mask older tool outputs~\citep{complexitytrap} or summarize
the history once it crosses a token threshold, as Claude
Code~\citep{claudecode} and Codex~\citep{codex} do; prompt compressors keep the
tokens that information scores or learned predictors rate as
useful~\citep{selectivecontext,llmlingua,longllmlingua,llmlingua2,xiao2025verbalized}. Such
criteria track size or salience, which need not coincide with the turn at which
a piece of evidence stops being needed. Tool-based methods let the agent decide
what to edit and when, through tools that fragment, summarize, hide, and
restore its history~\citep{sculptor} or fold completed sub-trajectories into
summaries~\citep{contextfolding}. The harness defines these tools, which ties
the learned policy to that harness. 
NCE represents each edit with the model's 
vocabulary (\S\ref{sec:editable_state}) and trains on edits that modify a block
only after release analysis finds it no longer needed (\S\ref{sec:edit_data}).

\paragraph{Learning model-native capabilities.}
Native multimodal models learn images inside one model, as shared-vocabulary
tokens or raw patches trained jointly with
text~\citep{chameleon,emu3,eve,nmmscaling}; earlier
systems attach a separate encoder through an adapter~\citep{flamingo,llava}.
Nativeness thus keeps a capability in the model's own sequence and
parameters. By this standard, learned control is only partly native: tool
calls and folding directives draw meaning from external tools or
prompts~\citep{toolformer,agentfold}, reflection tokens trigger an external
retriever~\citep{selfrag}, and reduction tokens act only on
reasoning~\citep{pencil,memento}.
Following native multimodal models, NCE incorporates context editing directly into the model's vocabulary through operator and address tokens, which are processed and generated inline with text.

\section{Native Context Editing}
\label{sec:method}

\definecolor{nceRoleBg}{HTML}{E6E9ED}
\definecolor{nceRoleFg}{HTML}{5E6570}
\definecolor{nceInk}{HTML}{2B2F36}
\definecolor{nceMute}{HTML}{868D97}
\definecolor{nceRowRew}{HTML}{FDF0D6}
\definecolor{nceRowDrop}{HTML}{FBE6E5}
\definecolor{ncePanA}{HTML}{F4F5F7}
\definecolor{ncePanALine}{HTML}{DDE1E6}
\definecolor{ncePanAHead}{HTML}{3E4451}
\definecolor{ncePanB}{HTML}{EDF3FB}
\definecolor{ncePanBLine}{HTML}{CCDBEF}
\definecolor{ncePanBHead}{HTML}{24508A}
\definecolor{ncePanC}{HTML}{EEF6EF}
\definecolor{ncePanCLine}{HTML}{CBE3CF}
\definecolor{ncePanCHead}{HTML}{2C6A3B}

\tikzset{
  nce pill/.style={rounded corners=1.3pt, inner xsep=0.85pt, inner ysep=0pt,
    text height=4.1pt, text depth=1.25pt, font=\fontsize{4.9}{6}\selectfont\ttfamily},
}
\newcommand{\ncePill}[3]{\tikz[baseline=(ncep.base)]\node[nce pill,fill=#1,text=#2](ncep){#3};\hspace{0.45pt}}
\newcommand{\nceSid}[2]{\ncePill{nceAddrBg}{nceAddrFg}{<sid\_hi\_#1>}\ncePill{nceAddrBg}{nceAddrFg}{<sid\_lo\_#2>}}
\newcommand{\nceSidFaint}[2]{\ncePill{nceAddrBg!55}{nceAddrFg!60}{<sid\_hi\_#1>}\ncePill{nceAddrBg!55}{nceAddrFg!60}{<sid\_lo\_#2>}}
\newcommand{\nceSidGone}[2]{\tikz[baseline=(nceg.base)]{%
  \node[nce pill,fill=nceDropBg!70,text=nceDropFg!85](nceg){<sid\_hi\_#1>};
  \node[nce pill,fill=nceDropBg!70,text=nceDropFg!85,anchor=west](nceh) at ([xshift=0.45pt]nceg.east){<sid\_lo\_#2>};
  \draw[nceDropFg,line width=0.5pt] (nceg.west) -- (nceh.east);}\hspace{0.45pt}}
\newcommand{\nceCtx}{\ncePill{nceEnvBg}{white}{<ctx>}}
\newcommand{\nceCtxEnd}{\ncePill{nceEnvBg}{white}{</ctx>}}
\newcommand{\nceOpDrop}{\ncePill{nceDropBg}{nceDropFg}{<op\_drop>}}
\newcommand{\nceOpRewrite}{\ncePill{nceRewBg}{nceRewFg}{<op\_rewrite>}}
\newcommand{\nceRole}[1]{\ncePill{nceRoleBg}{nceRoleFg}{<#1>}}
\newcommand{\nceObs}{\nceRole{longcat\_observation}\nceRole{longcat\_tool\_response}}
\newcommand{\nceT}[1]{\leavevmode{\ttfamily\color{nceInk}#1}}
\newcommand{\nceN}[1]{\leavevmode{\rmfamily\itshape\color{nceMute}#1}}
\newcommand{\nceE}{\leavevmode{\color{nceMute}\kern0.9pt$\cdots$\kern0.9pt}}
\newcommand{\nceNl}{\leavevmode{\color{nceMute}\scalebox{0.62}[0.8]{$\hookleftarrow$}}}
\newcommand{\nceNum}[1]{\leavevmode{\ttfamily\color{nceMute}#1}}
\newcommand{\nceUnch}{\nceN{unchanged}}
\newcommand{\nceGoneTxt}{\nceN{\textcolor{nceDropFg!80}{removed}}}
\newcommand{\nceHead}[3]{{\fontsize{6.6}{7.5}\selectfont\rmfamily\color{#1}\textbf{(#2)}~#3}}

\newlength{\ncePad}\newlength{\nceGap}\newlength{\ncePanCW}\newlength{\ncePanAW}
\newlength{\nceAW}\newlength{\nceCW}\newlength{\nceDW}\newlength{\nceEW}
\setlength{\ncePad}{5pt}
\setlength{\nceGap}{4pt}
\setlength{\ncePanCW}{168pt}
\setlength{\ncePanAW}{\dimexpr\textwidth-\nceGap-\ncePanCW\relax}
\setlength{\nceCW}{20pt}
\setlength{\nceEW}{13pt}
\setlength{\nceAW}{\dimexpr\ncePanAW-2\ncePad-\nceCW\relax}
\setlength{\nceDW}{\dimexpr\ncePanCW-2\ncePad-\nceEW\relax}
\newcommand{\nceRow}[4]{%
  \parbox[t]{\nceAW}{\raggedright#1}%
  \parbox[t]{\nceCW}{\raggedleft\nceNum{#2}}%
  \hspace{\dimexpr2\ncePad+\nceGap\relax}%
  \parbox[t]{\nceDW}{\raggedright#3}%
  \parbox[t]{\nceEW}{\raggedleft#4}}

\begin{figure*}[t]
\centering
\fontsize{5.45}{6.6}\selectfont
\makebox[\textwidth][c]{\begin{tikzpicture}[
    row/.style={anchor=north west, inner xsep=0pt, inner ysep=1.05pt, text width=\textwidth-2\ncePad},
    hdr/.style={anchor=base west, inner sep=0pt},
  ]
  \coordinate (o) at (0,0);
  \node[hdr] at (\ncePad,-10pt) {\nceHead{ncePanAHead}{a}{Rendered context $\operatorname{Render}(S_t)$ }};
  \node[hdr] at (\ncePanAW+\nceGap+\ncePad,-10pt) {\nceHead{ncePanCHead}{c}{Blocks after execution}};

  \node[row] (r0) at (\ncePad,-13pt) {\nceRow
    {\nceSid{00}{00}\nceN{system prompt and tool declarations (locked)}}
    {14,905}{\nceSidFaint{00}{00}\nceUnch}{}};
  \node[row] (r1) at (r0.south west) {\nceRow
    {\nceSid{00}{01}\nceRole{longcat\_user}\nceE\nceT{"Endless Drift" when}\nceE\nceT{path to start with "/"}\nceE}
    {1,310}{\nceSidFaint{00}{01}\nceUnch}{}};
  \node[row] (r2) at (r1.south west) {\nceRow
    {\nceSid{00}{02}\nceN{to}\,\nceSid{00}{06}\nceN{exploration and file reads}}
    {7,020}{\nceSidFaint{00}{02}\nceN{to}\,\nceSidFaint{00}{06}}{}};
  \node[row] (r7) at (r2.south west) {\nceRow
    {\nceSid{00}{07}\nceObs\nceT{Grep}\nceNl\nceT{Found 3 files}\nceNl\nceE}
    {42}{\nceSidGone{00}{07}\nceGoneTxt}{\nceNum{\textcolor{nceDropFg}{0}}}};
  \node[row] (r11) at (r7.south west) {\nceRow
    {\nceSid{00}{11}\nceObs\nceT{Bash}\nceNl\\
     \nceT{=== RUN~~~TestUnitTemplateResource}\nceNl\nceT{===~RUN~~~TestUnitTemplateResource/Full\_GitLab}\nceE\\
     \nceT{=== PAUSE TestUnitTemplateResource/Full\_GitLab\_EE\_template\_(without\_SSH\_keys)}\nceNl\nceE}
    {2,019}{\nceSid{00}{11}\nceRole{longcat\_observation}\\
     \nceRole{longcat\_tool\_response}\nceT{Bash}\nceNl\nceT{Baseline test run}\\
     \nceT{(before any fix):}\nceE\nceT{PASS, ok}\nceE\nceT{9.633s.}\nceE}
    {\nceNum{\textcolor{nceRewFg}{122}}}};
  \node[row] (r12) at (r11.south west) {\nceRow
    {\nceSid{00}{12}\nceObs\nceT{Bash}\nceNl\nceE\nceT{PASS}\nceNl\nceT{ok}\nceE\nceT{9.633s}}
    {1,207}{\nceSidGone{00}{12}\nceGoneTxt}{\nceNum{\textcolor{nceDropFg}{0}}}};
  \node[row] (r13) at (r12.south west) {\nceRow
    {\nceSid{00}{13}\nceObs\nceT{Write}\nceNl\nceT{File created}\nceE}
    {26}{\nceSidFaint{00}{13}\nceUnch}{}};
  \node[row] (rh) at (r13.south west) {\nceRow
    {\nceE\nceN{earlier actions ($A_t$, not addressable)}}
    {2,571}{}{}};
  \node[row] (r14) at (rh.south west) {\nceRow
    {\nceSid{00}{14}\nceObs\nceT{Bash}\nceNl\\
     \nceT{Exit code 1}\nceE\nceT{expected "misc/null-resource", got "/misc/null-resource"}\nceE\nceT{FAIL}}
    {186}{\nceSidFaint{00}{14}\nceUnch}{}};
  \node[row] (rl) at (r14.south west) {\nceRow
    {\nceE\nceN{one more action, then block 15}}
    {276}{\nceSidFaint{00}{15}\nceUnch}{}};
  \node[row] (rf) at ([yshift=-1.5pt]rl.south west) {\nceRow
    {\hfill\nceN{context length}}
    {\textcolor{nceInk}{29,562}}{\hfill\nceN{change}\kern20pt}{\llap{\nceNum{\textcolor{nceInk}{-3,146}}}}};

  \path let \p1=(rf.south) in coordinate (pbot) at (0,\y1-2.5pt);
  \coordinate (xa) at (\ncePanAW,0);
  \coordinate (xc) at (\ncePanAW+\nceGap,0);
  \coordinate (xr) at (\textwidth,0);
  \begin{scope}[on background layer]
    \fill[ncePanA, rounded corners=3pt] (o) rectangle (xa |- pbot);
    \fill[ncePanC, rounded corners=3pt] (xc) rectangle (xr |- pbot);
    \foreach \r/\c in {r7/nceRowDrop, r11/nceRowRew, r12/nceRowDrop} {
      \fill[\c] (o |- \r.north) rectangle (xa |- \r.south);
      \fill[\c] (xc |- \r.north) rectangle (xr |- \r.south);
    }
    \fill[nceDropFg] (o |- r7.north) rectangle ([xshift=1.4pt]o |- r7.south);
    \fill[nceRewFg!80] (o |- r11.north) rectangle ([xshift=1.4pt]o |- r11.south);
    \fill[nceDropFg] (o |- r12.north) rectangle ([xshift=1.4pt]o |- r12.south);
    \draw[ncePanALine, line width=0.4pt] ([xshift=\ncePad]o |- rf.north) -- ([xshift=-\ncePad]xa |- rf.north);
    \draw[ncePanCLine, line width=0.4pt] ([xshift=\ncePad]xc |- rf.north) -- ([xshift=-\ncePad]xr |- rf.north);
    \draw[ncePanALine, line width=0.4pt, rounded corners=3pt] (o) rectangle (xa |- pbot);
    \draw[ncePanCLine, line width=0.4pt, rounded corners=3pt] (xc) rectangle (xr |- pbot);
  \end{scope}

  \node[anchor=north west, inner xsep=\ncePad, inner ysep=3pt, fill=ncePanB, draw=ncePanBLine, line width=0.4pt,
        rounded corners=3pt, text width=\textwidth-2\ncePad] (out) at ([yshift=-\nceGap]o |- pbot) {%
    \nceHead{ncePanBHead}{b}{Model output at turn $t$: edits $e_t$ and action $a_t$}\par\vspace{2pt}%
    \renewcommand{\arraystretch}{1.0}%
    \begin{tabular}{@{}>{\raggedleft\arraybackslash}p{36pt}@{\hspace{5pt}}>{\raggedright\arraybackslash}p{\dimexpr\textwidth-2\ncePad-36pt-5pt-5pt-\nceEW\relax}@{\hspace{5pt}}>{\raggedleft\arraybackslash}p{\nceEW}@{}}
      \nceN{rationale} &
      \nceRole{longcat\_assistant}\nceT{The diagnosis phase just finished \textemdash{} the leading-slash path bug is confirmed by a failing reproduction test \textemdash{} so I've tidied up the leftover search and baseline-test output}\nceE\nceT{keeping only the conclusions (}\nceE\nceT{). I'm moving on to implementing the fix.} & \nceNum{77} \\[1.2pt]
      \nceN{rewrite} &
      \nceCtx\nceOpRewrite\nceSid{00}{11}\nceT{ Baseline test run (before any fix):}\nceE\nceT{every existing subtest of TestUnitTemplateResource passed (}\nceE\nceT{). Final result: PASS, ok}\nceE\nceT{9.633s. This is the clean baseline the fix must not break. }\nceCtxEnd & \nceNum{116} \\[1.2pt]
      \nceN{drop, drop} &
      \nceCtx\nceOpDrop\nceSid{00}{12}\nceCtxEnd\hspace{6pt}\nceCtx\nceOpDrop\nceSid{00}{07}\nceCtxEnd & \nceNum{5+5} \\[1.2pt]
      \nceN{action} &
      \nceRole{longcat\_tool\_call}\nceT{Edit}\nceE\nceT{resource\_template.go}\nceE\nceT{DiffSuppressFunc:}\nceE\nceT{strings.TrimLeft(old, "/") == strings.TrimLeft(new, "/")}\nceE & \\
    \end{tabular}};
\end{tikzpicture}}
\caption{\textbf{An abridged excerpt of an NCE training trajectory on a Go repository.} (a)~The rendered context at turn~$t$; (b)~the model's output, a rationale followed by three edits and an action; (c)~the blocks after execution. More detailed examples can be found in Appendix~\ref{app:case}.}
\label{fig:grammar_example}
\end{figure*}
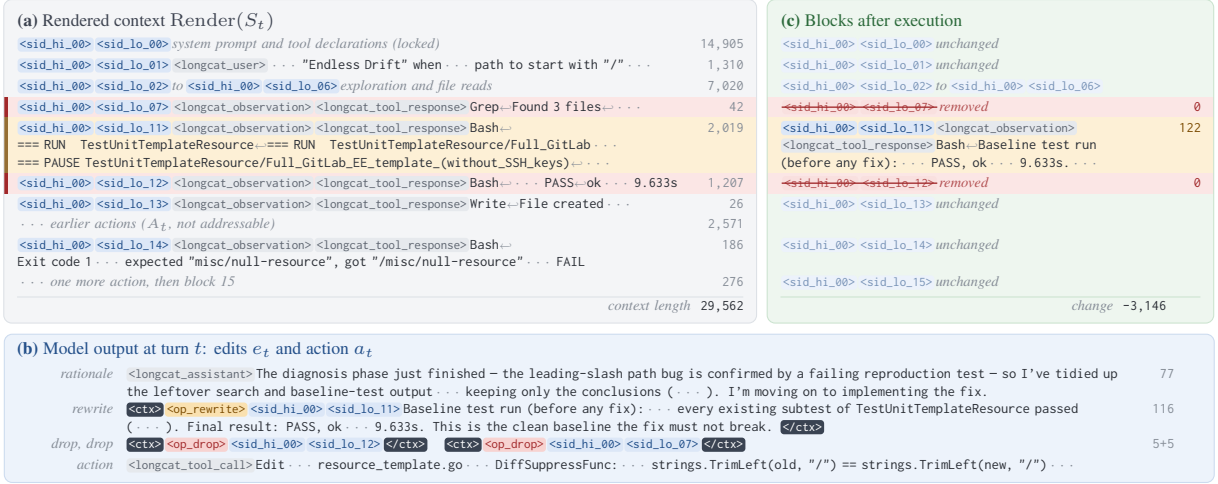

NCE lets an agent edit its own context by generating edits in an executable grammar (\S\ref{sec:editable_state}). Recorded trajectories contain no edits, so we derive them with two teachers (\S\ref{sec:edit_data}). Each edit changes what the agent reads next, so training follows the evolving context (\S\ref{sec:state_replay}).

\subsection{A Native Grammar for Context Editing}
\label{sec:editable_state}

\paragraph{Notations.}
An edit must name its target, so NCE first organizes the context into addressable blocks. At turn $t$, a block $b_t=(\alpha,x)$ pairs a persistent address $\alpha$ with text $x$, which may include system prompts, task descriptions, observations, or agent-written notes. The agent's action at turn $t$, such as a tool call, is denoted by $a_t$.
Context editing at turn $t$ is conditioned on the current state $S_t=(B_t,A_t)$, where the block history $B_t=[b_1,\ldots,b_{t-1}]$ contains the preceding context blocks, and the action history $A_t=[a_1,\ldots,a_{t-1}]$ records the agent's earlier actions. 
The model reads $\operatorname{Render}(S_t)$, which interleaves the blocks and actions in interaction order, with each block text prefixed by its address, and generates zero or more edits $e_t$ (Figure~\ref{fig:grammar_example}). When the turn ends, a task-agnostic runtime executes the edits and updates the histories with the current block $b_t$ and action $a_t$:
\begin{equation}
    S_{t+1}=\operatorname{Execute}(S_t,e_t)\oplus(a_t,b_t),
    \label{eq:nce_step}
\end{equation}
where $\oplus$ appends $a_t$ to the action history $A_t$ and $b_t$ to the edited block history $B_t$. With $e_t$ empty, Eq.~\eqref{eq:nce_step} is the ordinary growth of an agent's context. 

\paragraph{Vocabulary.}
NCE reserves 70 dedicated tokens. Each block address is encoded as a two-token pair \tokaddr{$h$}{$\ell$}, drawn from 32 high-order and 32 low-order address tokens. These 64 tokens support 1,024 distinct addresses. With addresses numbered from zero, the 72nd block is represented by \tokaddr{$02$}{$07$}. The model targets a block by copying its two address tokens. The remaining six tokens comprise three operators (\opdrop, \oprewrite, \opappend), two envelope delimiters (\tokctx{} and \tokctxend{}), and a marker for locked blocks.

\paragraph{Composition rules.}
An edit is one envelope around one operator:
\begin{equation*}
\begin{aligned}
e \;::=\;\; & \tokctx\;\opdrop\;\alpha_1\cdots\alpha_m\;\tokctxend\\
\mid\;\; & \tokctx\;\oprewrite\;\alpha\;x\;\tokctxend\\
\mid\;\; & \tokctx\;\opappend\;x\;\tokctxend
\end{aligned}
\end{equation*}
where each $\alpha$ is a block address, $m\ge1$, and $x$ is text represented by ordinary tokens. One turn may carry several edits, explained by a one-sentence rationale in plain text.

\paragraph{Semantics.}
Each edit denotes exactly one transition of the block history. \opdrop{} removes blocks $\alpha_1,\ldots,\alpha_m$ and retires their addresses for good, \oprewrite{} replaces the text of block $\alpha$ with $x$ while preserving its address, and \opappend{} adds an editable block with text $x$ under a fresh address. $\operatorname{Execute}$ checks addresses and permissions and applies the edits in $e_t$ in order. The edit instructions themselves are not retained in the context. $\operatorname{Render}(S_{t+1})$ contains the updated blocks and action history, and its KV cache is computed from that rendered text alone, indicating that dropped blocks are excluded from subsequent rendering and cache computation.

\begin{figure*}[t]
  \centering
  \includegraphics[width=1\textwidth]{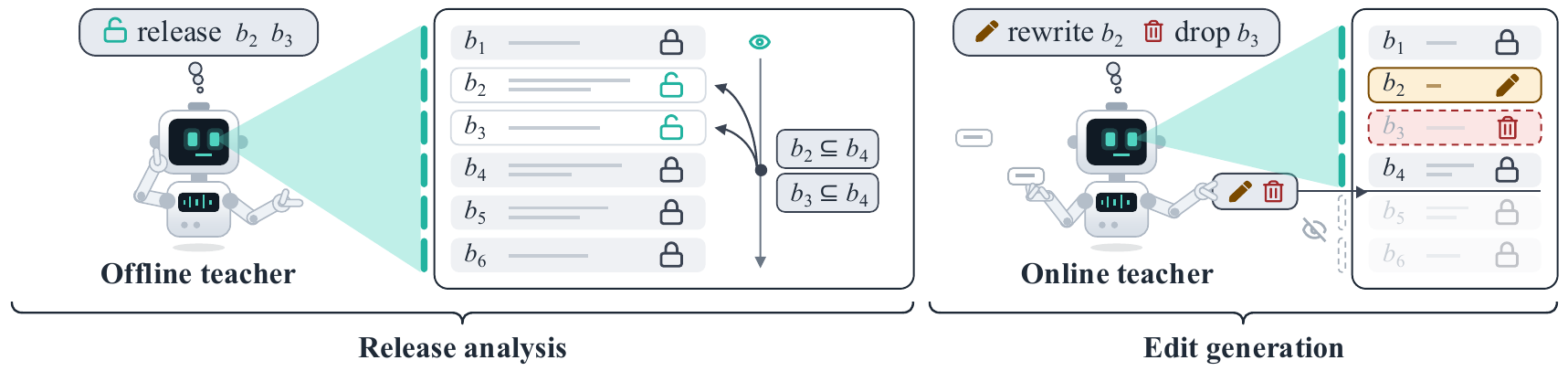}
 \caption{\textbf{Editing supervision from two teachers.} Left: the offline teacher reads the whole trajectory and releases a block only by citing the event that ended its use ($b_i\subseteq b_j$: $b_j$ supersedes or consumes $b_i$). Right: the online teacher reads only up to $b_4$ and edits released blocks, here rewriting $b_2$ and dropping $b_3$.}
  \label{fig:method}
\end{figure*}

\subsection{Faithful Supervision from Two Teachers}
\label{sec:edit_data}

We derive editing supervision from recorded agent trajectories, which contain no edits. A trajectory $\tau$ runs for $T$ turns from a task, and at turn $t$ the agent takes action $a_t$ and receives observation $b_t$. Replaying $\tau$ through Eq.~\eqref{eq:nce_step} with every $e_t$ empty yields its states $S_1,\ldots,S_T$, in which the task and each observation form a block. The central question is how long each block remains needed. A block $b$ is needed until the last turn whose action uses it,
\begin{equation}
    r_b=\max\,\{\,t : a_t \text{ uses } b\,\},
    \label{eq:nce_last_use}
\end{equation}
and $r_b$ is the turn at which $b$ enters if no action uses it. An edit to $b$ after turn $r_b$ is safe because no later action reads $b$, and delaying it further only keeps $b$ in context.

\paragraph{Offline teacher: release analysis.}
The offline teacher reads the whole trajectory and assigns each block a release point $\hat r_b$, the turn after which the block may be edited (Figure~\ref{fig:method}, left). Release analysis keeps every block by default and releases one only when the teacher cites the event that ended its use, typically an action that consumed the block or an observation that superseded it. Let $k_b$ denote the turn of the cited event, with $k_b=T$ if no event is cited. A mechanical check then traces the rare literals of the block, such as paths, hashes, and line numbers, through later actions. We use $\lambda_b$ to denote the last turn at which one of them reappears. The release point is
\begin{equation}
    \hat r_b=\max(k_b,\lambda_b).
    \label{eq:nce_release}
\end{equation}
A reappearing literal counts as a use, so $\lambda_b\le r_b$. The literal check can only postpone release; premature release ($\hat r_b<r_b$) therefore requires both an incorrect release judgment and a final use that leaves no literal trace. An omitted citation merely prolongs retention. Directly labeling $r_b$ requires identifying the exact last use, with errors leading to either premature release or unnecessary retention. Release analysis instead defaults to retention when evidence is missing, much as conservative liveness analysis keeps a value live while it may still be read. 
A second pass over the complete trajectory revisits blocks retained during the first pass, including those from abandoned lines of investigation, and releases them only with a cited event and after the same literal check.

\paragraph{Online teacher: edit generation.}
The online teacher decides when and how to edit the released blocks, seeing the present blocks and block-level eligibility markers derived from offline release analysis  (Figure~\ref{fig:method}, right). It acts as the agent at turn $t$: it reads $\operatorname{Render}(S_t)$, in which a block becomes editable only after its release point $\hat r_b$ from the offline teacher, and writes the edits $e_t$, choosing the operators and the replacement text. Every summary is drawn from the context in view, so no edit relies on anything the agent has yet to see. By default, it rewrites and drops only blocks that carry no information. Decision points arise when a reminder signals that the context is nearing its budget or when a subgoal has just been completed, and the teacher may decline when little is editable. A final audit rejects edits whose removed literals reappear later and summaries that state facts absent from their source.

\begin{figure*}[t]
  \centering
  \includegraphics[width=1\textwidth]{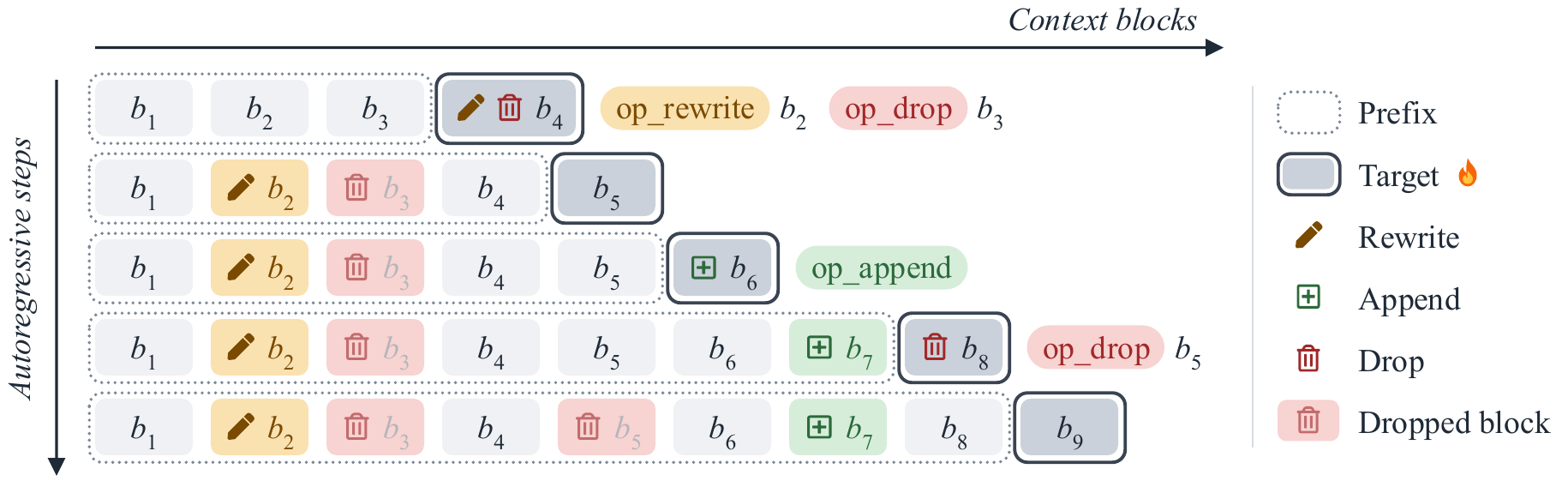}
  \caption{\textbf{Distillation with evolving contexts.} Each row supervises one turn (solid box) on the prefix left by all earlier edits (dotted box). Each editing turn closes a training sequence, so the five rows form four sequences (rows 1, 2--3, 4, and 5). Dropped blocks are drawn only for alignment.}
  \label{fig:train}
\end{figure*}

\subsection{Distillation with Evolving Contexts}
\label{sec:state_replay}

An edit changes every later rendering, so training on the recorded history would condition later turns on context the agent has already modified. We therefore build each training trajectory by running Eq.~\eqref{eq:nce_step} forward along $\tau$: $a_t$ and $b_t$ are the recorded action and observation, and $e_t$ is the online teacher's output at a decision point and empty elsewhere. Each editing decision is thus made on a context shaped by the earlier ones, as the decisions of the trained agent will be, and a summary written at one decision point is what the teacher reads at the next, where it may be revised or dropped in turn. Release analysis keeps the recorded actions valid on this edited context, since an edit removes only what the recorded continuation no longer uses.

Every turn with edits closes a segment, and the next segment opens with the rendering of the state those edits produce. Each segment is an ordinary causal sequence whose prefix reflects every earlier edit, so the three editing turns in Figure~\ref{fig:train} yield four training sequences. We fine-tune on these sequences with the usual next-token loss on the model's outputs, which covers actions, rationales, and edits alike; rendered prefixes and observations serve only as conditioning. Stretches where the teacher declined to edit, including whole trajectories, are trained in the same way, so the model learns when to leave its context alone as well as how to edit it.

\paragraph{Inference.}
Inference runs Eq.~\eqref{eq:nce_step} with the model writing $e_t$ and $a_t$ and the environment returning $b_t$: at each turn the model reads $\operatorname{Render}(S_t)$ and generates its output, and the runtime executes the edits and appends the action and its observation.

\section{Experiments and Results}
\label{sec:experiments}

\providecommand{\pend}[1][XX]{\textcolor{red}{[#1]}}

\definecolor{mrOurs}{HTML}{EDF3FB}
\definecolor{mrGain}{HTML}{2C6A3B}
\definecolor{mrMute}{HTML}{868D97}

\newcommand{\mrPara}[1]{{\color{mrMute}\textsc{#1}}}
\newcommand{\mrUp}{{\color{mrMute}\,$\uparrow$}}
\newcommand{\mrDown}{{\color{mrMute}\,$\downarrow$}}
\newcommand{\mrEst}{\rlap{$^{\dagger}$}}
\newcommand{\mrHead}[1]{\multicolumn{1}{c}{#1}}
\newcommand{\mrNCE}[3][2.1em]{#2\,\makebox[#1][l]{\scriptsize #3}}
\newcommand{\mrUpBy}[1]{{\color{mrGain}+#1}}
\newcommand{\mrDownBy}[1]{{\color{mrGain}$-$#1\%}}
\newcommand{\mrFlat}[1]{{\color{mrMute}#1}}

\begin{table*}[t]
\centering
\small
\setlength{\tabcolsep}{4.2pt}
\renewcommand{\arraystretch}{1.12}
\begin{tabular}{@{}l rr>{\columncolor{mrOurs}}l @{\hspace{10pt}} rr>{\columncolor{mrOurs}}l @{\hspace{16pt}} r>{\columncolor{mrOurs}}l @{\hspace{10pt}} r>{\columncolor{mrOurs}}l@{}}
\toprule
& \multicolumn{6}{c}{\textit{LongCat-Flash-Lite-Sparse (69B-A3B)}} & \multicolumn{4}{c}{\textit{Qwen3.6-35B-A3B}} \\
\cmidrule(lr{16pt}){2-7}\cmidrule(l){8-11}
& \multicolumn{3}{c}{Success (\%)\mrUp} & \multicolumn{3}{c}{Tokens ($\times10^6$)\mrDown} & \multicolumn{2}{c}{Success (\%)\mrUp} & \multicolumn{2}{c}{Tokens ($\times10^6$)\mrDown} \\
\cmidrule(lr{10pt}){2-4}\cmidrule(lr{16pt}){5-7}\cmidrule(lr{10pt}){8-9}\cmidrule(l){10-11}
\textbf{Benchmark} & \mrPara{rules} & \mrPara{tools} & \mrHead{\mrPara{native}} & \mrPara{rules} & \mrPara{tools} & \mrHead{\mrPara{native}} & \mrPara{rules} & \mrHead{\mrPara{native}} & \mrPara{rules} & \mrHead{\mrPara{native}} \\
& Comp. & Tools & \mrHead{\textbf{NCE}} & Comp. & Tools & \mrHead{\textbf{NCE}} & Comp. & \mrHead{\textbf{NCE}} & Comp. & \mrHead{\textbf{NCE}} \\
\midrule
TB 2.0         & 32.6 & 32.6 & \mrNCE[1.6em]{\textbf{38.2}}{\mrUpBy{5.6}} & 6.62 & 6.38  & \mrNCE{\textbf{4.60}}{\mrDownBy{31}} & 38.2 & \mrNCE[1.6em]{\textbf{43.8}}{\mrUpBy{5.6}} & \textbf{2.98} & \mrNCE{3.04}{\mrFlat{+2\%}} \\
SWE Verified   & 65.2 & 62.4 & \mrNCE[1.6em]{\textbf{70.0}}{\mrUpBy{4.8}} & 4.07       & 3.40  & \mrNCE{\textbf{2.70}}{\mrDownBy{34}} & 69.8 & \mrNCE[1.6em]{\textbf{74.0}}{\mrUpBy{4.2}} & 2.92 & \mrNCE{\textbf{2.19}}{\mrDownBy{25}} \\
SWE Multiling. & 56.0 & 54.3 & \mrNCE[1.6em]{\textbf{62.3}}{\mrUpBy{6.3}} & 10.83      & 10.93 & \mrNCE{\textbf{6.12}}{\mrDownBy{44}} & 58.0 & \mrNCE[1.6em]{\textbf{64.3}}{\mrUpBy{6.3}} & 5.12 & \mrNCE{\textbf{4.13}}{\mrDownBy{19}} \\
SWE Pro        & 39.4 & 39.7 & \mrNCE[1.6em]{\textbf{44.9}}{\mrUpBy{5.2}} & 9.50 & 8.46  & \mrNCE{\textbf{6.42}}{\mrDownBy{32}} & 43.6 & \mrNCE[1.6em]{\textbf{45.3}}{\mrUpBy{1.7}} & 5.82 & \mrNCE{\textbf{4.92}}{\mrDownBy{15}} \\
\midrule
\textit{Average} & 48.3 & 47.3 & \mrNCE[1.6em]{\textbf{53.9}}{\mrUpBy{5.6}} & 7.76     & 7.29  & \mrNCE{\textbf{4.96}}{\mrDownBy{36}} & 52.4 & \mrNCE[1.6em]{\textbf{56.9}}{\mrUpBy{4.5}} & 4.21 & \mrNCE{\textbf{3.57}}{\mrDownBy{15}} \\
\bottomrule
\end{tabular}
\caption{\textbf{Results at a 256K budget for harness compaction (Comp.), tool-call editing (Tools), and NCE}; TB~2.0 is Terminal-Bench 2.0, SWE is SWE-bench, and tokens are per task. Bold marks the best per base model. Small numbers give the success gain of NCE over the best baseline and its token change relative to harness compaction.}
\label{tab:main_results}
\end{table*}

\subsection{Experimental Setup}
\label{sec:setup}

\paragraph{Benchmarks.}
We evaluate on Terminal-Bench 2.0~\citep{terminalbench} (89 tasks), SWE-bench Verified~\citep{swebench} (500 issues), SWE-bench Multilingual~\citep{swemultilingual} (300 issues), and SWE-bench Pro~\citep{swepro} (731 issues in the public set). Terminal-Bench runs in Terminus-2, its reference agent, and the SWE-bench variants run in Claude Code~\citep{claudecode}. Neither harness is aware of NCE: the runtime sits between the harness and the model server, executes edits there, and returns ordinary responses, so a single checkpoint serves both harnesses without any editing interface. We report two metrics: the success rate of a single run, counting runs that end without a result as failures, and token usage, the prompt and output tokens of all model calls in a task, including compaction, averaged over tasks after excluding the 10\% with the highest and the 10\% with the lowest usage. Such a trimmed mean is used to avoid the influence of long-tail data. 
All four benchmarks are evaluated with a 256K-token budget. Terminal-Bench accumulates much longer histories than SWE-bench, so we also evaluate it at 128K, 64K, and 48K.

\paragraph{Models and training.}
We train NCE on LongCat-Flash-Lite-Sparse (69B-A3B)~\citep{longcatsparse} and Qwen3.6-35B-A3B~\citep{qwen36}, two mixture-of-experts models with 3B active parameters, starting from their instruction-tuned checkpoints.
We use GLM 5.3~\citep{glm5.3} as the two teachers of \S\ref{sec:edit_data} to annotate nearly 180K recorded agent trajectories on software engineering and terminal tasks, which we split into training segments (\S\ref{sec:state_replay}). Each model is fully fine-tuned on these segments for three epochs. Appendix~\ref{app:setup} details data, hyperparameters, and the budget configuration.

\paragraph{Baselines.}
We compare NCE with one baseline from each family of the Introduction, marked \textsc{rules} and \textsc{tools} in the tables.
\emph{Harness compaction} (Comp.) runs the base model that NCE is trained from, with the built-in compaction of each harness, which replaces the history with a model-written summary once it reaches the budget.
\emph{Tool-call editing} (Tools) is our implementation of tool-based editing following MemAct~\citep{memact}. We train it on LongCat only, using the same trajectories, edit supervision, and training recipe as NCE. Each edit invokes one of three tools that mirror NCE's operators (Appendix~\ref{app:setup}), enabling a controlled comparison between tool-call and native editing.
Our evaluation focuses on the active context under a fixed budget, so systems that store and retrieve information outside it fall outside this comparison.

\subsection{Main Results}
\label{sec:main_results}

\begin{table}[t]
\centering
\small
\setlength{\tabcolsep}{2.4pt}
\renewcommand{\arraystretch}{1.12}
\begin{tabular}{@{}l@{\hspace{7pt}} rr>{\columncolor{mrOurs}}l @{\hspace{7pt}} rr>{\columncolor{mrOurs}}l@{}}
\toprule
& \multicolumn{3}{c}{Success (\%)\mrUp} & \multicolumn{3}{c}{Tokens ($\times10^6$)\mrDown} \\
\cmidrule(lr{7pt}){2-4}\cmidrule(l){5-7}
\textbf{Budget} & Comp. & Tools & \mrHead{\textbf{NCE}} & Comp. & Tools & \mrHead{\textbf{NCE}} \\
\midrule
256K & 32.6 & 32.6 & \mrNCE[1.6em]{\textbf{38.2}}{\mrUpBy{5.6}} & 6.62 & 6.38 & \mrNCE{\textbf{4.60}}{\mrDownBy{31}} \\
128K & 25.8 & 30.3 & \mrNCE[1.6em]{\textbf{34.8}}{\mrUpBy{4.5}} & 6.08       & 4.96 & \mrNCE{\textbf{3.70}}{\mrDownBy{39}} \\
64K  & 27.0 & 29.2 & \mrNCE[1.6em]{\textbf{32.6}}{\mrUpBy{3.4}} & 3.42       & 3.58 & \mrNCE{\textbf{2.52}}{\mrDownBy{26}} \\
48K  & 20.2 & 27.0 & \mrNCE[1.6em]{\textbf{30.3}}{\mrUpBy{3.3}} & 3.40       & 3.17 & \mrNCE{\textbf{2.15}}{\mrDownBy{37}} \\
\bottomrule
\end{tabular}
\caption{\textbf{LongCat on Terminal-Bench 2.0 as the context budget shrinks}.}
\label{tab:tb_budgets}
\end{table}

\begin{figure*}[t]
  \centering
  \includegraphics[width=\textwidth]{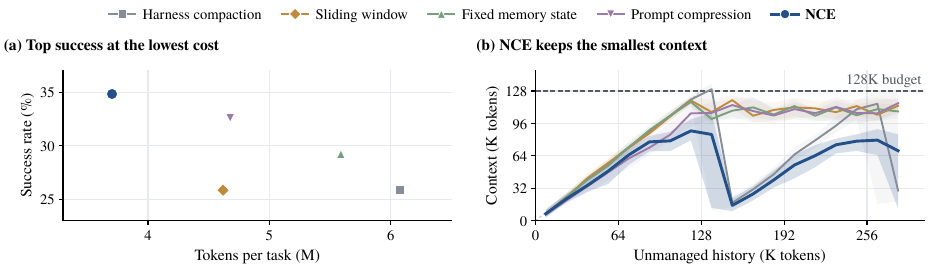}
  \caption{\textbf{Context management on Terminal-Bench 2.0 with LongCat under a 128K budget.} (a)~Success rate vs.\ tokens per task. (b)~Context length vs.\ unmanaged history length (median and interquartile range); the drop of NCE at 128K is the harness compaction it keeps.}
  \label{fig:context_mgmt}
\end{figure*}

\begin{figure*}[t]
  \centering
  \includegraphics[width=\textwidth]{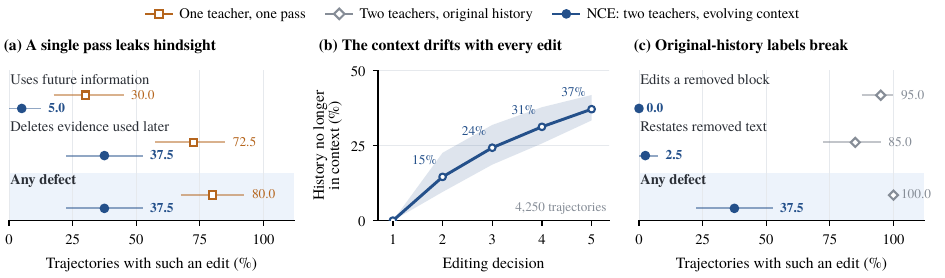}
  \caption{\textbf{Building editing supervision.} (a, c)~Share of the 40 trajectories with at least one edit of each kind (95\% bootstrap intervals). (b)~Share of the original history no longer in the agent's context at each editing decision, over 4,250 training trajectories (median and interquartile range).}
  \label{fig:supervision}
\end{figure*}

\paragraph{NCE achieves higher success with fewer tokens.}
At 256K, NCE leads both baselines on every benchmark while spending 36\% fewer tokens per task than harness compaction on average (Table~\ref{tab:main_results}). The gains hold in Claude Code and Terminus-2 alike, neither of which exposes an editing interface.

\paragraph{Native edits outperform the same edits as tool calls.}
Trained on the same edits, tool-call editing falls below harness compaction in average success and saves only 6\% of its tokens, while NCE saves 36\%. The edits pay off only when the model writes them in its own grammar.

\paragraph{The gains hold at every budget.}
On Terminal-Bench, NCE leads both baselines at every budget from 256K down to 48K, improving success over harness compaction by 5.6 to 10.1 points with 26\% to 39\% fewer tokens, and NCE at 64K matches both baselines at 256K with less than 40\% of their tokens (Table~\ref{tab:tb_budgets} and Figure~\ref{fig:main}b).

\paragraph{The gains transfer across base models.}
On Qwen3.6, NCE improves all four benchmarks and cuts average token usage by 15\%: the same grammar and supervision teach a second model family to edit its own context.

\subsection{Ablation Study}
\label{sec:ablation}

We compare NCE with other context managers, ablate how its supervision is built, and examine its edits at inference.

\paragraph{Native editing reaches the best success at the lowest cost.}
On Terminal-Bench under a 128K budget, we replace NCE with four context managers on the base model: harness compaction, a sliding window over recent turns, a fixed 4K memory state, and prompt compression of old tool outputs (Figure~\ref{fig:context_mgmt}, Appendix~\ref{app:ctx_mgmt}). NCE reaches the highest success rate (34.8\% against 25.8--32.6\%) with the fewest tokens per task (3.7M against 4.6--6.1M) and the smallest context (30.5K against 38--44K on average). The other four managers act on position, size, or the budget alone, while our NCE edits by what the task still needs, so its context stays below theirs throughout the task.

\paragraph{Supervision needs both teachers and the evolving context.}
We build supervision for the same 40 trajectories in three ways: by one teacher in one pass, by the two teachers on the original history, and by the two teachers on the evolving context, as in NCE. We check every edit against what the agent could see when it was made (Figure~\ref{fig:supervision}, Appendix~\ref{app:supervision}). Labeling in one pass leaks hindsight: in 30.0\% of trajectories a summary uses future information. Labeling on the original history misses that earlier edits have removed part of it, a median of 37\% by the fifth decision, so in 95.0\% of trajectories an edit targets a block that is already gone. NCE lowers these defects to 5.0\% and 0.0\%, and the share of trajectories with any defective edit from 80.0\% (one pass) and 100.0\% (original history) to 37.5\%.

\definecolor{opDropM}{HTML}{C64B48}
\definecolor{opRewM}{HTML}{DFA032}
\definecolor{opAppM}{HTML}{3A8F55}
\definecolor{nceInk}{HTML}{2B2F36}
\definecolor{nceMute}{HTML}{868D97}
\definecolor{nceGrid}{HTML}{E6E9ED}
\definecolor{nceLink}{HTML}{CDD2D9}
\providecommand{\nceFigFont}{\fontsize{7}{8.4}\selectfont}
\providecommand{\nceFigTitle}{\fontsize{7.5}{9}\selectfont\bfseries}
\providecommand{\nceFigNum}{\fontsize{6.5}{7.5}\selectfont}
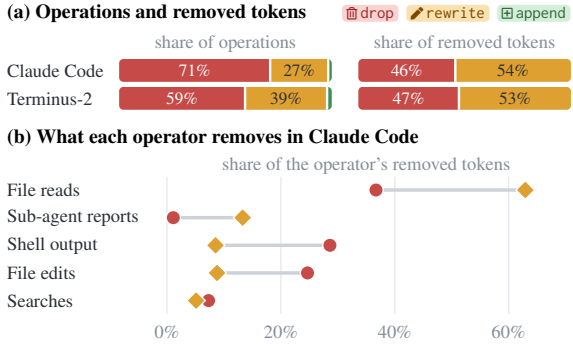
\begin{figure}[t]
\centering
\begin{tikzpicture}[x=1pt, y=1pt, font=\nceFigFont]
\node[anchor=west, font=\nceFigTitle] at (0,0) {(a) Operations and removed tokens};
\node[anchor=east, inner sep=0pt] at (216,0) {\opdrop\hspace{3pt}\oprewrite\hspace{3pt}\opappend};
\node[nceMute, anchor=base] at (86.0,-13) {share of operations};
\node[nceMute, anchor=base] at (176.0,-13) {share of removed tokens};
\node[anchor=west, nceInk] at (0,-21.5) {Claude Code};
\begin{scope}
\clip[rounded corners=1.6pt] (46,-25.80) rectangle (126,-17.20);
\fill[opDropM] (46.00,-25.80) rectangle (102.82,-17.20);
\fill[opRewM] (102.82,-25.80) rectangle (124.66,-17.20);
\fill[opAppM] (124.66,-25.80) rectangle (126.00,-17.20);
\draw[white, line width=0.9pt] (102.82,-26.80) -- (102.82,-16.20);
\draw[white, line width=0.9pt] (124.66,-26.80) -- (124.66,-16.20);
\end{scope}
\node[white, font=\nceFigNum] at (74.41,-21.50) {71\%};
\node[nceInk, font=\nceFigNum] at (113.74,-21.50) {27\%};
\begin{scope}
\clip[rounded corners=1.6pt] (136,-25.80) rectangle (216,-17.20);
\fill[opDropM] (136.00,-25.80) rectangle (172.49,-17.20);
\fill[opRewM] (172.49,-25.80) rectangle (216.00,-17.20);
\draw[white, line width=0.9pt] (172.49,-26.80) -- (172.49,-16.20);
\end{scope}
\node[white, font=\nceFigNum] at (154.25,-21.50) {46\%};
\node[nceInk, font=\nceFigNum] at (194.25,-21.50) {54\%};
\node[anchor=west, nceInk] at (0,-32) {Terminus-2};
\begin{scope}
\clip[rounded corners=1.6pt] (46,-36.30) rectangle (126,-27.70);
\fill[opDropM] (46.00,-36.30) rectangle (93.38,-27.70);
\fill[opRewM] (93.38,-36.30) rectangle (124.25,-27.70);
\fill[opAppM] (124.25,-36.30) rectangle (126.00,-27.70);
\draw[white, line width=0.9pt] (93.38,-37.30) -- (93.38,-26.70);
\draw[white, line width=0.9pt] (124.25,-37.30) -- (124.25,-26.70);
\end{scope}
\node[white, font=\nceFigNum] at (69.69,-32.00) {59\%};
\node[nceInk, font=\nceFigNum] at (108.81,-32.00) {39\%};
\begin{scope}
\clip[rounded corners=1.6pt] (136,-36.30) rectangle (216,-27.70);
\fill[opDropM] (136.00,-36.30) rectangle (173.86,-27.70);
\fill[opRewM] (173.86,-36.30) rectangle (216.00,-27.70);
\draw[white, line width=0.9pt] (173.86,-37.30) -- (173.86,-26.70);
\end{scope}
\node[white, font=\nceFigNum] at (154.93,-32.00) {47\%};
\node[nceInk, font=\nceFigNum] at (194.93,-32.00) {53\%};
\node[anchor=west, font=\nceFigTitle] at (0,-47) {(b) What each operator removes in Claude Code};
\node[nceMute, anchor=base] at (139.00,-59.00) {share of the operator's removed tokens};
\draw[nceGrid, line width=0.5pt] (64.00,-61.50) -- (64.00,-113.10);
\node[nceMute, anchor=north] at (64.00,-113.60) {0\%};
\draw[nceGrid, line width=0.5pt] (106.86,-61.50) -- (106.86,-113.10);
\node[nceMute, anchor=north] at (106.86,-113.60) {20\%};
\draw[nceGrid, line width=0.5pt] (149.71,-61.50) -- (149.71,-113.10);
\node[nceMute, anchor=north] at (149.71,-113.60) {40\%};
\draw[nceGrid, line width=0.5pt] (192.57,-61.50) -- (192.57,-113.10);
\node[nceMute, anchor=north] at (192.57,-113.60) {60\%};
\node[anchor=west, nceInk] at (0,-66.50) {File reads};
\draw[nceLink, line width=1.3pt, line cap=round] (142.64,-66.50) -- (198.83,-66.50);
\filldraw[fill=opDropM, draw=white, line width=0.5pt] (142.64,-66.50) circle (2.7pt);
\filldraw[fill=opRewM, draw=white, line width=0.5pt] (198.83,-63.00) -- (202.33,-66.50) -- (198.83,-70.00) -- (195.33,-66.50) -- cycle;
\node[anchor=west, nceInk] at (0,-76.90) {Sub-agent reports};
\draw[nceLink, line width=1.3pt, line cap=round] (66.38,-76.90) -- (92.47,-76.90);
\filldraw[fill=opDropM, draw=white, line width=0.5pt] (66.38,-76.90) circle (2.7pt);
\filldraw[fill=opRewM, draw=white, line width=0.5pt] (92.47,-73.40) -- (95.97,-76.90) -- (92.47,-80.40) -- (88.97,-76.90) -- cycle;
\node[anchor=west, nceInk] at (0,-87.30) {Shell output};
\draw[nceLink, line width=1.3pt, line cap=round] (125.31,-87.30) -- (82.28,-87.30);
\filldraw[fill=opDropM, draw=white, line width=0.5pt] (125.31,-87.30) circle (2.7pt);
\filldraw[fill=opRewM, draw=white, line width=0.5pt] (82.28,-83.80) -- (85.78,-87.30) -- (82.28,-90.80) -- (78.78,-87.30) -- cycle;
\node[anchor=west, nceInk] at (0,-97.70) {File edits};
\draw[nceLink, line width=1.3pt, line cap=round] (116.88,-97.70) -- (82.84,-97.70);
\filldraw[fill=opDropM, draw=white, line width=0.5pt] (116.88,-97.70) circle (2.7pt);
\filldraw[fill=opRewM, draw=white, line width=0.5pt] (82.84,-94.20) -- (86.34,-97.70) -- (82.84,-101.20) -- (79.34,-97.70) -- cycle;
\node[anchor=west, nceInk] at (0,-108.10) {Searches};
\draw[nceLink, line width=1.3pt, line cap=round] (79.65,-108.10) -- (74.96,-108.10);
\filldraw[fill=opDropM, draw=white, line width=0.5pt] (79.65,-108.10) circle (2.7pt);
\filldraw[fill=opRewM, draw=white, line width=0.5pt] (74.96,-104.60) -- (78.46,-108.10) -- (74.96,-111.60) -- (71.46,-108.10) -- cycle;
\end{tikzpicture}
\caption{\textbf{Operator use of NCE with LongCat at 256K.} (a)~Claude Code pools the three SWE-bench sets, and Terminus-2 runs Terminal-Bench 2.0; \opappend{} removes no tokens. (b)~Each operator's removed tokens by the tool that produced the block.}
\label{fig:operators}
\end{figure}

\paragraph{NCE drops what it has used and condenses what it still needs.}
\opdrop{} accounts for most operations, yet \oprewrite{} removes slightly more tokens (Figure~\ref{fig:operators}): it condenses long blocks, mostly file reads and sub-agent reports, to about a fifth of their length, while \opdrop{} also clears shell output and edit results. Both harnesses show the same pattern, and 99.1\% of the 73K edits NCE writes execute as written. Most edits are proactive: 89\% come before any budget reminder. Editing also frees budget early: when the unedited history first fills a 48K budget, NCE's context fills only 73\% of it, leaving over a quarter of the budget free (Appendix~\ref{app:inference}).

\paragraph{Re-prefill does not slow the agent.}
An edit invalidates the KV cache from the first edited block onward. Without any overlap, re-prefilling takes only 0.2--0.4\% of a task's wall-clock time. Since a turn writes its edits before its action, the runtime can start the re-prefill as soon as the last edit closes, while the action is still being generated and executed; on measured timings, this hides it entirely in 98\% of editing turns under Terminus-2 and 81\% under Claude Code. In compute, the edits remove 4 to 22 times as many tokens from later prompts as the re-prefill adds (Appendix~\ref{app:prefill}).

\section{Conclusion}
\label{sec:conclusion}

We introduced Native Context Editing (NCE), which makes context editing part of the language an LLM agent speaks. 
A grammar of dedicated tokens binds each edit to one transition of an addressable context state, so the meaning of every edit ships with the model and holds in every harness. 
Two teachers supply causal edits, while training follows the context produced by earlier edits. 
Across benchmarks, base models, and harnesses, NCE improves task success with fewer tokens and outperforms the same edits issued as tool calls. 
Context management thereby moves from the harness into the model's own language.

\section*{Limitations}
\label{sec:limitations}

We evaluate NCE on software engineering and terminal tasks, where long tool outputs make context editing especially consequential; extending it to other long-horizon settings, such as web navigation and deep research, is a natural next step. Both base models are mixture-of-experts models with 3B active parameters, and scaling NCE to larger models remains to be explored. NCE learns to edit through supervised distillation, and reinforcement learning on task outcomes could further refine when and how the agent edits. Finally, this study manages the active context within a fixed budget, and combining NCE with external memory and retrieval is left for future work.

\section*{Ethical Considerations}
\label{sec:ethics}

This work trains LLM agents on recorded trajectories of software engineering and terminal tasks and evaluates them on public benchmarks. It involves no human subjects, and we collect no personal data. We use all models and datasets in accordance with their licenses and intended research use, and we do not foresee ethical concerns beyond those common to LLM agents.

\section*{LLM Usage}
\label{sec:llm_usage}

We used large language models (LLMs) only to check grammar and refine the language of this paper. The research ideas, analyses, results, and conclusions are entirely the authors' own.

\bibliography{references}

\setlength{\textfloatsep}{20pt plus 2pt minus 4pt}
\setlength{\dbltextfloatsep}{20pt plus 2pt minus 4pt}
\appendix

\newcommand{\appTocRow}[4]{%
  \parbox[t]{1.6em}{\hyperref[#4]{\colorbox{nceTakeFg}{\makebox[0.9em]{\color{white}\bfseries\sffamily #1}}}}%
  \parbox[t]{\dimexpr\linewidth-1.6em\relax}{\hyperref[#4]{\textbf{#2}}\enspace{\color{black!45}\dotfill}\enspace\pageref{#4}\\[-1pt]
    {\small\color{black!70}#3}}\par\vspace{5pt}}
\twocolumn[{%
\begin{center}
{\color{nceTakeFg}\rule{\textwidth}{0.8pt}}\\[8pt]
{\LARGE\bfseries Appendix}\\[6pt]
{\large Editing as Language: Teaching LLM Agents a Grammar for Native Context Editing}\\[7pt]
{\color{nceTakeFg}\rule{\textwidth}{0.8pt}}
\end{center}
\vspace{2pt}
\begin{tcolorbox}[colback=nceTakeBg, colframe=nceTakeFg, boxrule=0pt, leftrule=2pt, sharp corners,
  left=8pt, right=8pt, top=6pt, bottom=2pt, width=\textwidth]
{\color{nceTakeFg}\faIcon[regular]{map}~\textbf{Contents.}}\enspace
{\small The appendix follows the life of an edit: how the grammar defines it, how the teachers write it, how we train and evaluate on it, what it does at inference, and how it unfolds in complete tasks.}
\par\vspace{6pt}
\appTocRow{A}{The NCE Grammar and Runtime}{Dedicated tokens, rendering, execution of edits with a worked turn, reminders, and serving an unmodified harness.}{app:grammar}
\appTocRow{B}{Building the Editing Supervision}{Annotation pipeline, offline and online teacher prompts, an annotated trajectory, and data statistics.}{app:annotation}
\appTocRow{C}{Experimental Details}{Harnesses, context budgets, training data and hyperparameters, the tool-call baseline, and decoding.}{app:setup}
\appTocRow{D}{Analysis}{Context management, building supervision, editing behavior at inference, and the cost of re-prefill (\S\ref{sec:ablation}).}{app:analysis}
\appTocRow{E}{Case Studies}{NCE through a complete Terminal-Bench task under Terminus-2 and a SWE-bench task under Claude Code.}{app:case}
\end{tcolorbox}
\vspace{12pt}
}]
\renewcommand{\topfraction}{0.85}\renewcommand{\dbltopfraction}{0.85}
\renewcommand{\textfraction}{0.12}\renewcommand{\floatpagefraction}{0.75}\renewcommand{\dblfloatpagefraction}{0.75}
\setcounter{topnumber}{3}\setcounter{dbltopnumber}{2}

\section{The NCE Grammar and Runtime}
\label{app:grammar}

This appendix specifies the grammar of \S\ref{sec:editable_state} in full: the dedicated tokens (\ref{app:vocab}), how a context state is rendered (\ref{app:render}), how the runtime executes edits (\ref{app:execute}), the reminders it issues (\ref{app:reminders}), and how it serves an unmodified harness (\ref{app:runtime}).

\subsection{Vocabulary}
\label{app:vocab}

Table~\ref{tab:app_vocab} lists the 70 tokens NCE adds to the vocabulary of the base model. Block $n$, counted from zero in order of creation, receives the address \tokaddr{$h$}{$\ell$} with $h=\lfloor n/32\rfloor$ and $\ell=n \bmod 32$. An address is never reused within a session, so it names the same block for as long as the block lives, and a dropped block cannot be confused with a later one. The role tags that open each block (\texttt{<longcat\_user>}, \texttt{<longcat\_observation>}, and so on) are the existing tokens of the base chat template.

\begin{table}[h]
\centering
\footnotesize
\setlength{\tabcolsep}{3pt}
\begin{tabular}{@{}l c >{\raggedright\arraybackslash}p{0.36\columnwidth}@{}}
\toprule
\textbf{Token} & \textbf{\#} & \textbf{Role} \\
\midrule
\nceChip{nceAddrBg}{nceAddrFg}{<sid\_hi\_00>}\,\ldots\,\nceChip{nceAddrBg}{nceAddrFg}{<sid\_hi\_31>} & 32 & high address digit \\
\nceChip{nceAddrBg}{nceAddrFg}{<sid\_lo\_00>}\,\ldots\,\nceChip{nceAddrBg}{nceAddrFg}{<sid\_lo\_31>} & 32 & low address digit \\
\midrule
\opdrop & 1 & remove blocks \\
\oprewrite & 1 & replace a block's text \\
\opappend & 1 & add a note block \\
\midrule
\tokctx{}~\tokctxend & 2 & delimit one edit \\
\nceChip{nceEnvBg}{white}{<m\_ro>} & 1 & mark a locked block \\
\midrule
\textbf{Total} & \textbf{70} & \\
\bottomrule
\end{tabular}
\caption{\textbf{The dedicated tokens of NCE.} Operators are shown with the chips used throughout the paper; their token strings are \texttt{<op\_drop>}, \texttt{<op\_rewrite>}, and \texttt{<op\_append>}.}
\label{tab:app_vocab}
\end{table}

\subsection{Rendering}
\label{app:render}

$\operatorname{Render}(S_t)$ writes each block as its two address tokens, the lock marker if the block is locked, and the block text in the chat format of the base model, opened by its role tag. No closing token is needed: the address of the next block marks where a block ends. Actions are written as ordinary assistant turns and carry no address, so an edit can reach observations, the task, and notes, and leaves the agent's own actions intact. Every tool result forms a block of its own, which lets the agent drop one result of a batch and keep the others. Figure~\ref{fig:app_render} shows the beginning of a rendered context from a SWE-bench Multilingual task under Claude Code.

\begin{figure*}[t]
\centering
\begin{tcolorbox}[colback=white, colframe=nceEnvBg!55, boxrule=0.6pt, arc=2pt,
  left=6pt, right=6pt, top=4pt, bottom=4pt, width=\textwidth,
  title={\small\faIcon{stream}~~Rendered context at turn 22 (abridged)}, fonttitle=\sffamily\bfseries,
  colbacktitle=nceEnvBg, coltitle=white]
\footnotesize
\renewcommand{\arraystretch}{1.25}
\begin{tabular}{@{}p{2.8cm} >{\raggedright\arraybackslash}p{\dimexpr\linewidth-2.8cm-2\tabcolsep\relax}@{}}
\tokaddr{00}{00}\,\nceChip{nceEnvBg}{white}{<m\_ro>} & \texttt{\color{gray}<longcat\_tool\_declare>}\,\texttt{\# Tools You have access to the following tools: [\ldots]} \hfill{\color{gray}\scriptsize system prompt and tool declarations, 16.5K tokens, locked} \\
\tokaddr{00}{01} & \texttt{\color{gray}<longcat\_user>}\,\texttt{I have uploaded a c code repository in the /testbed directory. Now consider the following Github issue: [BUG] CLUSTER SHARDS command returns "empty array" in slots section [\ldots]} \hfill{\color{gray}\scriptsize task} \\
 & \texttt{\color{gray}<longcat\_assistant>}\,{\color{nceEnvBg!80}\itshape action: Task (delegates the root-cause analysis to a sub-agent)} \\
\tokaddr{00}{02} & \texttt{\color{gray}<longcat\_observation><longcat\_tool\_response>}\,\texttt{Task [\{"type":"text","text":"I now have all the information needed. Here is my complete analysis. [\ldots]} \\
 & \texttt{\color{gray}<longcat\_assistant>}\,{\color{nceEnvBg!80}\itshape action: Read src/cluster\_legacy.c} \\
\tokaddr{00}{03} & \texttt{\color{gray}<longcat\_observation><longcat\_tool\_response>}\,\texttt{Read 5380→ \} else if (server.cluster->importing\_slots\_from[j]) \{ [\ldots]} \\
 & \multicolumn{1}{c}{\color{gray}$\vdots$} \\
\tokaddr{00}{22} & \texttt{\color{gray}<longcat\_observation><longcat\_tool\_response>}\,\texttt{Task [\{"type":"text","text":"The survey of the cluster test suite [\ldots]} \\
\end{tabular}
\end{tcolorbox}
\caption{\textbf{The beginning of a rendered context.} Each block opens with its address; the system prompt carries the lock marker. Actions, set in italics here, are assistant turns without an address. Contents are abridged with [\ldots].}
\label{fig:app_render}
\end{figure*}

\subsection{Execution}
\label{app:execute}

When a turn ends, the runtime extracts every envelope \tokctx\ldots\tokctxend{} from the output and executes the edits in the order they were written. An edit executes only if it is well formed: the envelope opens with exactly one operator, \opdrop{} lists at least one address, \oprewrite{} names one address followed by its replacement text, and \opappend{} carries non-empty text. Every address must name a live block, and no edit may touch the locked block. An edit that fails these checks changes nothing, with one exception: a rewrite of a block that is no longer live is applied as an append, so that its text is kept. An edit repeated verbatim within a turn executes once. The runtime then strips the envelopes from the output, and the remaining text, the rationale and the action, enters the action history. If at least one edit executed, the next prompt is rendered afresh from the new state; otherwise it extends the previous prompt and reuses its prefix cache.

\begin{figure*}[t]
\centering
\begin{minipage}[t]{0.405\textwidth}
\vspace{0pt}
\begin{tcolorbox}[colback=white, colframe=nceEnvBg!55, boxrule=0.6pt, arc=2pt,
  left=5pt, right=5pt, top=4pt, bottom=4pt,
  title={\small\faIcon{robot}~~Model output at turn 22 (abridged)}, fonttitle=\sffamily\bfseries,
  colbacktitle=nceEnvBg, coltitle=white]
\scriptsize\raggedright
{\itshape The root-cause analysis and fix for CLUSTER SHARDS are done, so I've cleaned out the exploration snippets and intermediate edit snapshots from that phase, keeping the key file/line references and the final fix as it now stands in cluster\_legacy.c. Next up: writing and running the regression test.}

\smallskip
\tokctx\,\oprewrite\,\tokaddr{00}{03}~Exploration snippet (src/cluster\_legacy.c, \textasciitilde lines 5388--5459) read during root-cause analysis: clusterGenNodesSlotsInfo(int filter) at line 5392 populates slot\_info\_pairs [\ldots] kept only as line-number reference.\,\tokctxend

\smallskip
\tokctx\,\oprewrite\,\tokaddr{00}{05}~Exploration grep results: cluster.h declares clusterNodeGetPrimary at src/cluster.h:92; [\ldots] Not relevant to the fix; kept only as a one-line reference.\,\tokctxend

\smallskip
\tokctx\,\oprewrite\,\tokaddr{00}{06}~Pre-fix copy of addShardReplyForClusterShards (src/cluster\_legacy.c, lines 5665--5693): it fetched the primary via clusterNodeGetPrimary(n) and, if the primary's slot\_info\_pairs was NULL, emitted an empty slots array (addReplyArrayLen(c, 0)) --- the bug. Superseded by the current fixed implementation of the same function.\,\tokctxend

\smallskip
\tokctx\,\opdrop\,\tokaddr{00}{10}\,\tokctxend

\smallskip
\tokctx\,\oprewrite\,\tokaddr{00}{07}~slot\_info\_pairs API locations (from grep): struct fields at src/cluster\_legacy.h:299--300 [\ldots] clusterGenNodesSlotsInfo is called from clusterCommandShards at 5700.\,\tokctxend

\smallskip
\tokctx\,\opdrop\,\tokaddr{00}{11}\,\tokctxend\ \ {\color{gray}\ldots\ eight more \opdrop{} edits (blocks 12--18, 21)}

\smallskip
Now let me look at the existing test file to understand how the tests are structured [\ldots]\\
{\color{nceEnvBg!80}\itshape action: Read tests/cluster/tests/28-cluster-shards.tcl}
\end{tcolorbox}
\end{minipage}\hfill
\begin{minipage}[t]{0.575\textwidth}
\vspace{0pt}
\begin{tcolorbox}[colback=white, colframe=nceEnvBg!55, boxrule=0.6pt, arc=2pt,
  left=5pt, right=4pt, top=4pt, bottom=4pt,
  title={\small\faIcon{layer-group}~~Blocks before and after the turn (tokens)}, fonttitle=\sffamily\bfseries,
  colbacktitle=nceEnvBg, coltitle=white]
\centering
\scriptsize
\begin{tikzpicture}[x=1cm,y=1cm]
\begin{scope}[xshift=0cm]
\input{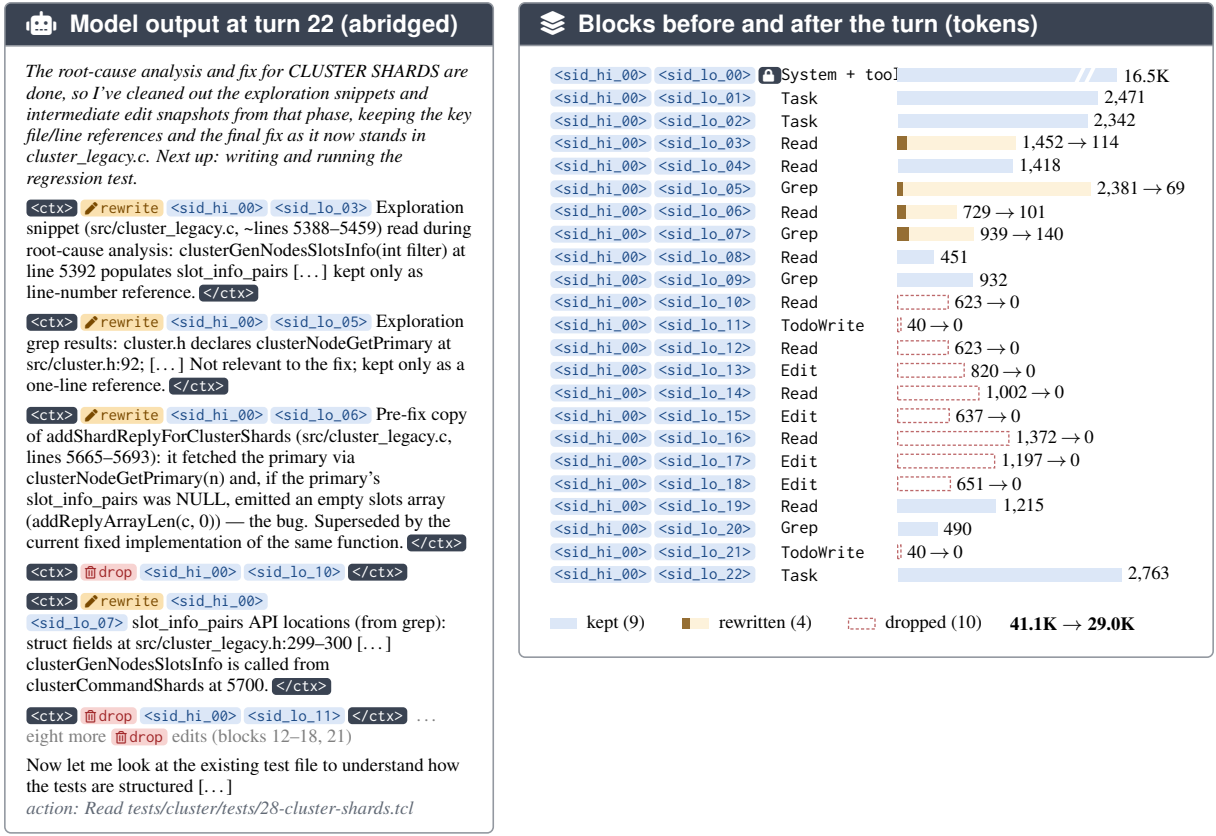}
\end{scope}
\begin{scope}[yshift=-7.25cm]
\fill[nceAddrBg] (0,-0.07) rectangle (0.35,0.07); \node[anchor=west,font=\scriptsize] at (0.4,0) {kept (9)};
\fill[nceRewBg!45] (1.75,-0.07) rectangle (2.1,0.07); \fill[nceRewFg!80] (1.75,-0.07) rectangle (1.85,0.07);
\node[anchor=west,font=\scriptsize] at (2.15,0) {rewritten (4)};
\draw[nceDropFg!70,dash pattern=on 1.2pt off 0.8pt,line width=0.5pt] (3.95,-0.07) rectangle (4.3,0.07);
\node[anchor=west,font=\scriptsize] at (4.35,0) {dropped (10)};
\node[anchor=west,font=\scriptsize\bfseries] at (6.0,0) {41.1K $\to$ 29.0K};
\end{scope}
\end{tikzpicture}
\end{tcolorbox}
\end{minipage}
\caption{\textbf{One editing turn, executed.} Left: the output of NCE at turn 22 of a SWE-bench Multilingual task under Claude Code, with no reminder pending; the edits are typeset with the chips of Table~\ref{tab:app_vocab}. Right: every block of the context before the turn and its size after it. Blocks 10, 12, 14, and 16 are earlier reads of the function that block 19 shows in its fixed form; blocks 13, 15, 17, and 18 confirm intermediate edits to it. Rewritten blocks keep their addresses; dropped addresses are retired.}
\label{fig:app_exec}
\end{figure*}

Figure~\ref{fig:app_exec} traces one editing turn from the task of Figure~\ref{fig:app_render}. After four rounds of reading and editing the same function, the agent has fixed the bug and turns to the regression test. Without a reminder, it writes a one-sentence rationale and fourteen edits before its next action. It drops the four superseded reads of the function and the four confirmations of its intermediate edits, keeps the latest read, which shows the fixed code, and rewrites four exploration results into one-line references that keep their file paths and line numbers. The context shrinks from 41.1K to 29.0K tokens, and the agent continues with the test.

\subsection{Reminders}
\label{app:reminders}

Under a budget $W$, the runtime appends a transient reminder to the prompt once the context reaches 70\% of $W$ (soft) or 95\% of $W$ (hard). The text contains no numbers, so the model cannot read off how much room is left, and the same text serves NCE and tool-call editing:
\begin{tcolorbox}[colback=nceRewBg!30, colframe=nceRewFg!60, boxrule=0.4pt, arc=2pt, left=5pt, right=5pt, top=3pt, bottom=3pt,
  fontupper=\scriptsize\ttfamily, title={\scriptsize\sffamily\bfseries\faBell[regular]~~Soft reminder (70\% of $W$)}, colbacktitle=nceRewBg!80, coltitle=nceRewFg]
[context-monitor] Your context has grown sizeable; this is a good moment to tidy it up. If anything in it is no longer needed, condense or remove all of it in one pass --- clear out everything that has served its purpose and leave the rest untouched; do not trim a little and save the rest for later. If there is very little worth cleaning up, or everything is still in use, you may explain why and decline --- that is a reasonable choice.
\end{tcolorbox}
\begin{tcolorbox}[colback=nceDropBg!35, colframe=nceDropFg!60, boxrule=0.4pt, arc=2pt, left=5pt, right=5pt, top=3pt, bottom=3pt,
  fontupper=\scriptsize\ttfamily, title={\scriptsize\sffamily\bfseries\faBell~~Hard reminder (95\% of $W$)}, colbacktitle=nceDropBg!90, coltitle=nceDropFg]
[context-monitor] Your context is close to its limit and must be cleaned up now. Before anything else this turn, condense or remove everything that is no longer needed --- clear all of it out and leave what you still rely on untouched. This is not optional; only if there is genuinely nothing that can be cleaned up may you explain why and skip it.
\end{tcolorbox}
A soft reminder may be declined. After any reminder the runtime waits four turns before the next one, eight after a declined soft reminder, and it stops issuing soft reminders once two have been declined; hard reminders are not limited. A reminder stays in the prompt only until the next edit executes. The online teacher sees reminders of the same form when it builds the supervision (Appendix~\ref{app:annotation}), and at evaluation most edits are issued without any reminder (Table~\ref{tab:inference_behavior}).

\subsection{Serving an Unmodified Harness}
\label{app:runtime}

The runtime runs as a proxy between the harness and the model server and keeps one context state per agent session, including each sub-agent that Claude Code launches. For every chat request it (i)~matches the request to its session and adds the new observations as blocks, (ii)~appends a reminder if one is due, (iii)~renders the state and forwards it to the model server, (iv)~executes the edits in the response, and (v)~returns the response to the harness with the envelopes removed and the action in the format of the harness. The harness therefore sees an ordinary model: its prompts, tool definitions, and compaction logic stay unchanged. Requests outside the agent loop, such as the title requests of Claude Code, pass through to the base chat template. Because the runtime holds the state, the same checkpoint serves Claude Code and Terminus-2 without any harness-specific code beyond request parsing.

\section{Building the Editing Supervision}
\label{app:annotation}

\newtcolorbox{ncePrompt}[1]{colback=nceAddrBg!25, colframe=nceEnvBg, coltitle=white,
  fonttitle=\footnotesize\bfseries, title={#1}, equal height group=appprompt, arc=1.5pt, boxrule=0.6pt,
  left=3pt, right=3pt, top=2pt, bottom=2pt, toptitle=1pt, bottomtitle=1pt,
  fontupper=\fontsize{6.4}{7.6}\selectfont\ttfamily\raggedright}
\newcommand{\ncePh}[1]{\colorbox{nceRewBg}{\textcolor{nceRewFg}{\{\{#1\}\}}}}
\newcommand{\nceCut}{\textcolor{gray}{[...]}}
\newcommand{\nceHd}[1]{\textbf{\textcolor{nceAddrFg}{#1}}}

\begin{figure*}[t]
\centering
\begin{tikzpicture}[
  stage/.style={draw=#1, fill=#1!12, rounded corners=2pt, line width=0.6pt,
    text width=2.12cm, minimum height=1.8cm, align=center, inner sep=2.5pt,
    font=\scriptsize},
  arr/.style={-{Stealth[length=4pt]}, line width=0.7pt, color=nceEnvBg}]
\node[stage=gray] (s1) {\textbf{Recorded trajectory}\\[1pt] task, actions $a_t$, observations; no edits};
\node[stage=nceAddrFg, anchor=west] (s2) at ([xshift=0.36cm]s1.east) {\textbf{Blocks}\\[1pt] task and each observation get an address};
\node[stage=nceAddrFg, anchor=west] (s3) at ([xshift=0.36cm]s2.east) {\textbf{Offline teacher}\\[1pt] release analysis: cited event, literal check, second pass $\Rightarrow \hat r_b$};
\node[stage=nceRewFg, anchor=west] (s4) at ([xshift=0.36cm]s3.east) {\textbf{Online teacher}\\[1pt] at decision points on the evolving context $\Rightarrow e_t$};
\node[stage=nceDropFg, anchor=west] (s5) at ([xshift=0.36cm]s4.east) {\textbf{Final audit}\\[1pt] literal reuse, unsupported facts, edit density};
\node[stage=nceAppFg, anchor=west] (s6) at ([xshift=0.36cm]s5.east) {\textbf{Segments}\\[1pt] one causal sequence per editing turn};
\foreach \a/\b in {s1/s2, s2/s3, s3/s4, s4/s5, s5/s6} \draw[arr] (\a) -- (\b);
\end{tikzpicture}
\caption{\textbf{The annotation pipeline.} Only the online teacher writes edits; the offline teacher decides which blocks it may touch, and the audit removes edits that fail a mechanical check.}
\label{fig:app_pipeline}
\end{figure*}
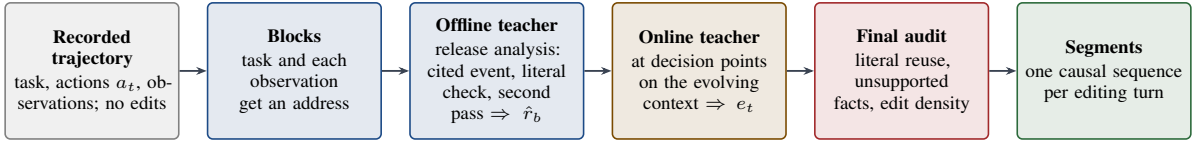

Figure~\ref{fig:app_pipeline} summarizes how a recorded trajectory becomes training segments. This section gives the rules behind each stage, the two teacher prompts, an annotated example, and statistics of the resulting data.

\subsection{Pipeline}

\paragraph{Blocks.}
We replay each trajectory with every $e_t$ empty and split it into blocks: the system prompt (locked), the task, every tool result, and every user message each form one block, and the block that enters before turn $t$ has entry turn $t$. Reasoning and actions stay in the action history.

\paragraph{Offline teacher, forward pass.}
The teacher reads the whole trajectory, with each block labeled by its index and entry turn, and keeps every block by default. It releases a block only by citing the event that ended its use: an action that consumed the block, such as writing the file it was read from into new code, or a later observation that superseded it, such as a full read of a file that a search result only pointed to. Each release names the turn after which the block may be edited and states its cause in one sentence that names the turn and the action. A release has one of two strengths: \emph{to summary}, when the fact that the block existed and its conclusion still help navigation, and \emph{to drop}, when the block carries no information, such as an operation receipt or an empty search result. The forward pass also marks at most two subgoal milestones per trajectory, which later become decision points.

\paragraph{Literal check.}
For each released block, we extract its rare literals: paths of at least four characters, identifiers containing an underscore, integers of five or more digits, hexadecimal strings of eight or more characters, decimals with four or more fractional digits, and four-digit numbers in line-number position, such as \texttt{1589:} in a search result. A literal is rare if it is at least four characters long and occurs at most eight times in the whole trajectory. We then scan every action after the cited turn, including tool-call arguments. If a literal reappears, the release moves to the last turn where one reappears, and a drop becomes a summary. Identifiers are traced only for drops, since a summary keeps them. Two further clamps only postpone as well: a block read from a file is not released before the last turn that edits that file, and a block that serves as the blueprint of the final solution is not released before the turn that writes the solution out.

\begin{figure*}[t]
\centering
\begin{minipage}[t]{0.495\textwidth}\vspace{0pt}
\begin{ncePrompt}{\faIcon{eye}~Offline teacher: release analysis (forward pass)}
\nceHd{\#\# SYSTEM}\\
You are a "context release adjudicator". The input is one completed agent trajectory, composed of INGEST blocks (numbered by block\_id, coming from the environment/tools/user) and GEN turns (numbered by turn, model outputs). You have a full-hindsight view. \nceCut\\[2pt]
The single red line is faithfulness: information that later text still needs must never be compressed away. So by default every block is kept, and only when you explicitly confirm that the details of a block will never be used again do you emit a release event for it.\\[2pt]
> Core asymmetry: missing a release event = the block is kept a little longer, which is safe; emitting a release event by mistake = later text loses information, a serious violation. \nceCut\\[2pt]
\nceHd{\#\#\# Task one: release-event identification}\\
- `to\_gist' (rewrite): the verbatim details are no longer needed, but "it existed, and here is its conclusion" still has navigational value. Usually: already digested \nceCut; already superseded (a locating result whose target was later fetched in full) \nceCut\\
- `to\_drop': zero information increment from start to finish (a pure operation acknowledgement, an empty match, an unrelated dead end). \nceCut\\[2pt]
Every release event supplies block\_id, after\_turn (>= the block's appearance turn), kind, and trigger: the causal event, naming which turn's which action digested/superseded/locked in this block. If you cannot find such a clear-cut causal point, do not release. \nceCut\\[2pt]
\nceHd{\#\#\# Task two: sub-goal milestones}\\
At most 2 for the entire trajectory \nceCut\\[2pt]
\nceHd{\#\#\# Output format}\\
\{"skip": false, "releases": [\{"block\_id": <int>, "after\_turn": <int>, "kind": "to\_gist|to\_drop", "trigger": "..."\}], "milestones": [\{"after\_turn": <int>, "description": "..."\}]\}\\[2pt]
\nceHd{\#\# USER}\\
\ncePh{meta}\\ \ncePh{trajectory: blocks with appearance turns, GEN turns}
\end{ncePrompt}
\end{minipage}\hfill
\begin{minipage}[t]{0.495\textwidth}\vspace{0pt}
\begin{ncePrompt}{\faPen~Online teacher: edit generation (decision point)}
\nceHd{\#\# SYSTEM}\\
You are now playing the agent itself, in the middle of executing a task, at turn \ncePh{turn}. You have just received a trigger signal and need to decide whether to edit your own context right now, and exactly how. You are the person on the scene, not an omniscient observer. \nceCut\\[2pt]
You can see the trigger signal and all surviving blocks, each with an "editable / off-limits" marker. Off-limits = still in use now or later (or locked); never touch it. \nceCut\\[2pt]
\nceHd{\#\#\# Available operations}\\
\{"op": "rewrite", "block": <id>, "text": "..."\}: the default first choice. Every fact in it must come from that block's original text. \nceCut\\
\{"op": "drop", "block": <id>\}: high bar, zero-information blocks only. \nceCut\\
\{"op": "append", "text": "..."\}: consolidates the shared conclusions of several blocks into one note. \nceCut\\[2pt]
\nceHd{\#\#\# Faithfulness hard limit}\\
After compressing, you can still finish the task with what remains. \nceCut\\[2pt]
\nceHd{\#\#\# Constant compression intensity}\\
Dispose of every block that is no longer needed; leave the blocks that are still needed exactly as they are. Either compress everything that should be compressed this time, or do not compress at all. \nceCut\\[2pt]
\nceHd{\#\#\# Decision rules}\\
Suggested tidy-up: if editable blocks are few, declining is encouraged; a token edit is worse than no edit. Mandatory tidy-up: no\_op only if not a single block is editable. Milestone: declinable. \nceCut\\[2pt]
\nceHd{\#\#\# Output format}\\
\{"no\_op": false, "note": "<one or two sentences, the visible rationale>", "edits": [...], "reason": "<internal, for reviewers>"\}\\[2pt]
\nceHd{\#\# USER}\\
\ncePh{trigger}\\ \ncePh{surviving blocks: id, source, editable marker, content}
\end{ncePrompt}
\end{minipage}
\caption{\textbf{Teacher prompts, abridged (\textcolor{gray}{[...]}).} Placeholders filled by the annotation driver are highlighted. The online teacher sees no turn after $t$, no release point, and no token count.}
\label{fig:app_prompts}
\end{figure*}

\paragraph{Offline teacher, second pass.}
The forward pass is conservative and keeps blocks such as the outputs of abandoned lines of investigation. A second pass collects candidates from the kept blocks in three ways: the teacher nominates blocks that, looking back over the finished trajectory, sit on an abandoned or superseded branch, with recall as the goal; a mechanical rule adds tool results whose rare literals stop reappearing at least three turns before the end; and injected reminders and bare continuation messages are added as well. A strict review then judges every candidate. A release must name the turn and action that ended the use of the block, and the review must find no residual dependency in a turn-by-turn scan of the rest of the trajectory, covering literals copied into later tool arguments (such as a line offset), later code checked against the block, and candidate options of an open decision. When the reviewer claims that a later block supersedes a candidate, we verify mechanically that the distinctive lines of the candidate appear in that block and reject the release otherwise. Approved releases pass the same literal check.

\paragraph{Decision points.}
Each trajectory is assigned a context budget $W$ from 32K, 48K, 64K, and 128K, the largest tier under which its peak context crosses the soft line while its incompressible floor still fits. Replaying the trajectory turn by turn with exact token counts, we create a decision point when (i) the context first reaches $0.70W$, which inserts a soft reminder that the agent may decline; (ii) it reaches $0.95W$, which inserts a hard reminder that it may decline only when nothing is editable; or (iii) a milestone of the forward pass is reached. Each trigger has a cool-down of two turns, and a milestone fires only if the last edit lies at least two turns back. At soft and milestone decision points, if the editable blocks together hold less than 8\% of the current context, the teacher is not called and the turn becomes a declined decision. The reminder text contains no numbers, and the teacher never sees token counts.

\begin{figure*}[t]
\centering
\footnotesize
\begin{tcolorbox}[colback=white, colframe=nceAddrFg, arc=1.5pt, boxrule=0.6pt, left=4pt, right=4pt, top=2pt, bottom=3pt,
  fonttitle=\footnotesize\bfseries, coltitle=white, title={\faIcon{eye}~(a) Offline teacher: release points}]
\setlength{\tabcolsep}{4pt}
\begin{tabular}{@{}l >{\raggedright\arraybackslash}p{4.5cm} >{\raggedright\arraybackslash}p{6.9cm} c c c@{}}
\toprule
Block & Content & Cited event (teacher, abridged) & $k_b$ & $\lambda_b$ & $\hat r_b$ \\
\midrule
5 & search hits for \texttt{validateCCVChannelParams} & Turn 4 read \texttt{provider/ibc\_module.go} in full; the hit list only pointed the way. & 5 & -- & 5 \\
9 & receipt: edit to \texttt{provider/keeper/proposal.go} & Pure success receipt; turn 11 verifies the change with a \texttt{Read}. & 8 & \cellcolor{nceRewBg}11 & \cellcolor{nceRewBg}11 \\
10 & receipt: edit to \texttt{consumer/ibc\_module.go} & Pure success receipt; turn 10 verifies the change with a \texttt{Read}. & 9 & \cellcolor{nceRewBg}10 & \cellcolor{nceRewBg}10 \\
14 & one unit test passes & Superseded by the full consumer test run at turn 14. & 13 & -- & 13 \\
\bottomrule
\end{tabular}\\[3pt]
\scriptsize Shaded: the literal check postpones the release, since the path in the receipt reappears in \texttt{Read(file\_path=...)} at the shaded turn; the teacher proposed a drop, which becomes a summary.
\end{tcolorbox}
\vspace{2pt}
\begin{tcolorbox}[colback=white, colframe=nceRewFg, arc=1.5pt, boxrule=0.6pt, left=4pt, right=4pt, top=2pt, bottom=3pt,
  fonttitle=\footnotesize\bfseries, coltitle=white, title={\faPen~(b) Online teacher: decision point after turn 12 (milestone: all three changes verified)}]
\raggedright
The three code changes are done and verified, so I condense the original file contents I read before the change and the edit receipts into key points, keep the conclusions, and move on to testing.\\[2pt]
\tokctx\,\oprewrite\,\tokaddr{00}{02}\,Read the original \texttt{x/ccv/consumer/ibc\_module.go} (package consumer, before the change). [...] \texttt{validateCCVChannelParams} (around lines 91--99) checks \texttt{portID == boundPort} (else \texttt{ErrInvalidPort}) and \texttt{version == ccv.Version} (else \texttt{ErrInvalidVersion}). [...]\,\tokctxend\\[1pt]
\tokctx\,\oprewrite\,\tokaddr{00}{09}\,Edit receipt: \texttt{/testbed/x/ccv/provider/keeper/proposal.go} updated successfully.\,\tokctxend\\[1pt]
\tokctx\,\oprewrite\,\tokaddr{00}{10}\,Edit receipt: \texttt{/testbed/x/ccv/consumer/ibc\_module.go}, first change applied.\,\tokctxend\\[1pt]
\textcolor{gray}{[three further rewrites: the second half of the original file read, the second receipt, and an earlier read of \texttt{proposal.go}]}
\end{tcolorbox}
\caption{\textbf{Annotation of one trajectory.} (a) Release points of four blocks; $k_b$ is the turn of the cited event and $\lambda_b$ the last turn at which a rare literal of the block reappears. (b) The online teacher's rationale and edits at a decision point, in the NCE grammar; block $n$ has address \tokaddr{00}{$n$}.}
\label{fig:app_example}
\end{figure*}

\paragraph{Online teacher.}
At a decision point the teacher plays the agent at turn $t$. It sees the trigger and every block still in context, each marked \emph{editable} if $t>\hat r_b$ and \emph{off-limits} otherwise; it does not see release points, release reasons, or anything after turn $t$. Editing is all or nothing: once the teacher edits, it disposes of every editable block, rewriting by default, dropping only blocks without information, and appending a note when several blocks share a conclusion that the note consolidates. Every fact in a summary must come from the block it replaces. The teacher writes the one-sentence rationale that precedes the edits in training. If the edits free less than 8\% of the context at a soft or milestone point, the whole decision is discarded and the turn is trained as a decline. The edits execute immediately, so the next decision point sees the rewritten blocks; a summary is editable again only if it is still long.

\paragraph{Final audit.}
The audit rejects an edit when (i) a number, path, identifier, or quoted string in a summary or note does not occur in its source block, or in the trajectory for a note; (ii) a literal that the edit removes reappears in a later action without having appeared in a summary or in visible output first; or (iii) the editing turn frees less than 3\% of the context. It discards a trajectory when a hard reminder ends without an edit, or when more than six turns, and more than 15\% of all turns, carry edits. About 4\% of annotated trajectories are discarded.

\paragraph{Segments.}
We render each surviving trajectory along the evolving context (\S\ref{sec:state_replay}) and cut a segment at every turn with edits. Reminders enter as transient user messages that are kept only in the segment in which they appear.

\subsection{Teacher Prompts}

Figure~\ref{fig:app_prompts} shows the prompt of the forward pass of the offline teacher and the prompt of the online teacher, abridged. The second pass uses two further prompts with the same output conventions: a nomination prompt that asks for high recall and a review prompt that enforces the release conditions above.

\subsection{An Annotated Example}

The trajectory fixes a channel-validation bug in a Go repository in 19 turns and 20 blocks (Figure~\ref{fig:app_example}); outputs are abridged and shown in English. The offline teacher releases two edit receipts at the turn after the edit, citing that they acknowledge an operation and nothing else. The literal check finds the edited path in a \texttt{Read} at a later turn and postpones both releases to that read, turning the drops into summaries. The online teacher acts at the milestone after turn 12, when all three code changes have been verified, and rewrites the original read of the edited file together with the receipts; each receipt summary keeps the path that the literal check traced.

\subsection{Data Statistics}

Table~\ref{tab:app_anno_stats} summarizes one annotation batch of about 50K trajectories. Decision points arise mostly at milestones and soft reminders; hard reminders are rare because edits at earlier decision points keep the context below $0.95W$. Edits make up close to a third of the supervised tokens, and close to a third of the segments carry no edit at all, which teaches the model when to leave its context alone.

\begin{table}[t]
\centering
\small
\begin{tabular}{@{}lr@{}}
\toprule
Statistic & Value \\
\midrule
Segments per trajectory & $\approx$2.9 \\
Decision points per trajectory & $\approx$2.0 \\
\quad milestone / soft / hard & 60 / 38 / 2\% \\
Segments without edits & 31\% \\
Edited blocks: rewrite / drop & 35 / 65\% \\
Appended notes, share of operations & 2\% \\
Supervised tokens that are edits & 31\% \\
Trajectories discarded by the audit & 4\% \\
\bottomrule
\end{tabular}
\caption{\textbf{Statistics of one annotation batch.} A drop that names several blocks counts once per block.}
\label{tab:app_anno_stats}
\end{table}

\section{Experimental Details}
\label{app:setup}

\begin{figure*}[t]
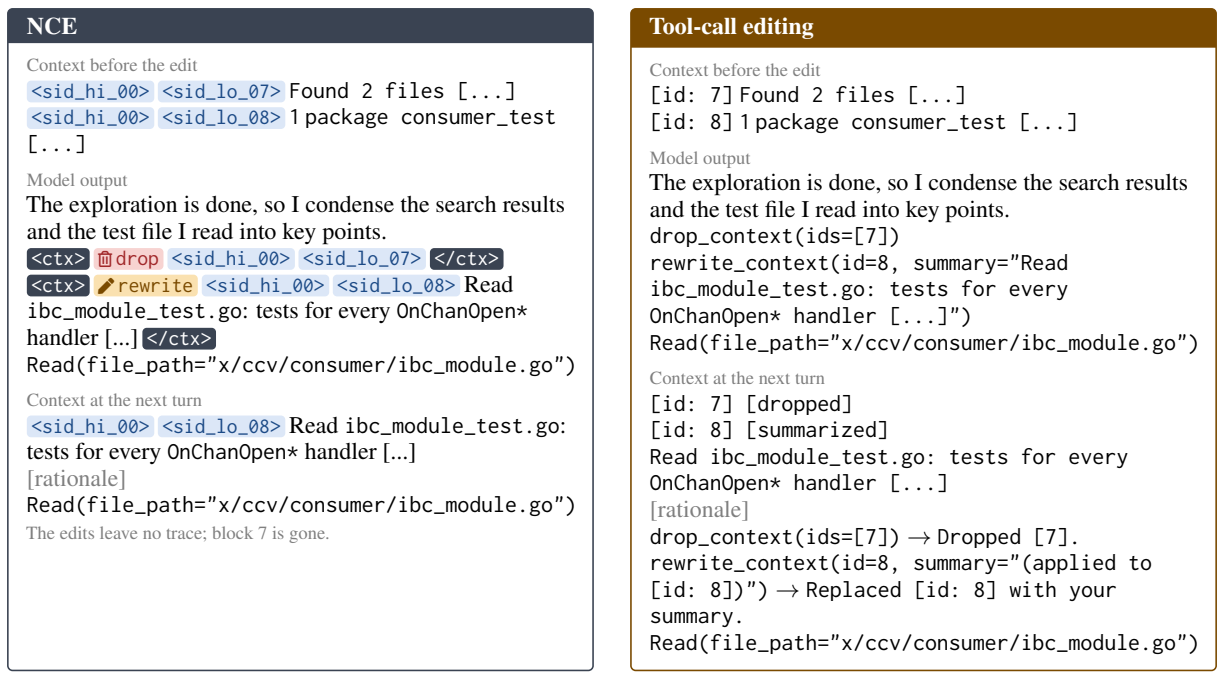

\centering
\footnotesize
\begin{minipage}[t]{0.485\textwidth}\vspace{0pt}
\begin{tcolorbox}[colback=white, colframe=nceEnvBg, arc=1.5pt, boxrule=0.6pt, left=4pt, right=4pt, top=3pt, bottom=3pt,
  fonttitle=\footnotesize\bfseries, coltitle=white, title={NCE}, equal height group=apptc]
\raggedright
\textcolor{gray}{\scriptsize Context before the edit}\\
\tokaddr{00}{07}\,\texttt{Found 2 files [...]}\\
\tokaddr{00}{08}\,\texttt{1\,package consumer\_test [...]}\\[3pt]
\textcolor{gray}{\scriptsize Model output}\\
The exploration is done, so I condense the search results and the test file I read into key points.\\
\tokctx\,\opdrop\,\tokaddr{00}{07}\,\tokctxend\\
\tokctx\,\oprewrite\,\tokaddr{00}{08}\,Read \texttt{ibc\_module\_test.go}: tests for every \texttt{OnChanOpen*} handler [...]\,\tokctxend\\
\texttt{Read(file\_path="x/ccv/consumer/ibc\_module.go")}\\[3pt]
\textcolor{gray}{\scriptsize Context at the next turn}\\
\tokaddr{00}{08}\,Read \texttt{ibc\_module\_test.go}: tests for every \texttt{OnChanOpen*} handler [...]\\
\textcolor{gray}{[rationale]}\\
\texttt{Read(file\_path="x/ccv/consumer/ibc\_module.go")}\\
\textcolor{gray}{\scriptsize The edits leave no trace; block 7 is gone.}
\end{tcolorbox}
\end{minipage}\hfill
\begin{minipage}[t]{0.485\textwidth}\vspace{0pt}
\begin{tcolorbox}[colback=white, colframe=nceRewFg, arc=1.5pt, boxrule=0.6pt, left=4pt, right=4pt, top=3pt, bottom=3pt,
  fonttitle=\footnotesize\bfseries, coltitle=white, title={Tool-call editing}, equal height group=apptc]
\raggedright
\textcolor{gray}{\scriptsize Context before the edit}\\
\texttt{[id: 7]} \texttt{Found 2 files [...]}\\
\texttt{[id: 8]} \texttt{1\,package consumer\_test [...]}\\[3pt]
\textcolor{gray}{\scriptsize Model output}\\
The exploration is done, so I condense the search results and the test file I read into key points.\\
\texttt{drop\_context(ids=[7])}\\
\texttt{rewrite\_context(id=8, summary="Read ibc\_module\_test.go: tests for every OnChanOpen* handler [...]")}\\
\texttt{Read(file\_path="x/ccv/consumer/ibc\_module.go")}\\[3pt]
\textcolor{gray}{\scriptsize Context at the next turn}\\
\texttt{[id: 7] [dropped]}\\
\texttt{[id: 8] [summarized]}\\ \texttt{Read ibc\_module\_test.go: tests for every OnChanOpen* handler [...]}\\
\textcolor{gray}{[rationale]}\\
\texttt{drop\_context(ids=[7])} $\rightarrow$ \texttt{Dropped [7].}\\
\texttt{rewrite\_context(id=8, summary="(applied to [id: 8])")} $\rightarrow$ \texttt{Replaced [id: 8] with your summary.}\\
\texttt{Read(file\_path="x/ccv/consumer/ibc\_module.go")}
\end{tcolorbox}
\end{minipage}
\caption{\textbf{The same two edits in NCE and in tool-call editing.} NCE edits vanish once executed, while each tool call stays in the history with its result message.}
\label{fig:app_toolcall}
\end{figure*}

\paragraph{Harnesses.}
The SWE-bench variants run in Claude Code (version 2.1.7) and Terminal-Bench 2.0 runs in Terminus-2 on Harbor. Both harnesses send ordinary chat requests to the NCE runtime, which serves them as described in Appendix~\ref{app:runtime}; neither harness needs any change to its prompts or tool definitions.

\paragraph{Context budgets.}
For a budget $W$, the compaction threshold of each harness is set to $W$: Claude Code through its auto-compaction setting and Terminus-2 through its context limit. Terminus-2 counts its own copy of the history, which keeps the text NCE has edited away, so it also compacts NCE runs once that copy reaches $W$. The runtime issues a soft reminder when the context reaches 70\% of $W$ and a hard reminder at 95\% (Appendix~\ref{app:reminders}); the reminder text contains no numbers and is identical for NCE and tool-call editing.

\paragraph{Training data.}
The training trajectories, nearly 180K in total, are downsampled from agentic data that LongCat has already been trained on. Appendix~\ref{app:annotation} describes how we annotate them with edits.

\paragraph{Training.}
LongCat is trained with Megatron on a large-scale GPU cluster. Qwen3.6 is trained with ms-swift using AdamW ($\beta_1=0.9$, $\beta_2=0.95$, weight decay $5\times10^{-5}$) at a peak learning rate of $1\times10^{-5}$, with 3\% linear warmup and cosine decay to $1\times10^{-6}$. Sequences are packed to 98,304 tokens, each global batch holds 128 packed sequences, and gradients are clipped to a norm of 1.0. Training runs in bfloat16 for three epochs (2,628 steps). For both models, the loss covers assistant tokens only.

\paragraph{Tool-call editing.}
The three tools take block identifiers shown in the rendered context: \texttt{drop\_context(ids)} removes blocks, \texttt{rewrite\_context(id, summary)} replaces a block in place, and \texttt{append\_note(note)} adds a note. Each call returns a result message, and a dropped or rewritten block leaves a short stub in the history. Every edit in the NCE training data is mapped to the corresponding tool call, so the two training sets contain the same edits at the same turns. Figure~\ref{fig:app_toolcall} shows one edit in both forms.

\begin{table*}[t]
\centering
\small
\setlength{\tabcolsep}{5pt}
\begin{tabular}{l cc c cc cc}
\toprule
 & & & & \multicolumn{2}{c}{\textbf{Context (K)}} & & \\
\cmidrule(lr){5-6}
\textbf{Method} & \textbf{Solved} & \textbf{Success (\%)} & \textbf{Tokens (M)} & \textbf{Mean} & \textbf{Peak} & \textbf{Reductions} & \textbf{Turns} \\
\midrule
Harness compaction & 23 & 25.8 & 6.08 & 38.3 & 79.7 & 0.6 & 119 \\
Sliding window & 23 & 25.8 & 4.62 & 43.6 & 71.5 & 1.7 & 92 \\
Fixed memory state & 26 & 29.2 & 5.59 & 44.0 & 72.9 & 1.9 & 95 \\
Prompt compression & 29 & 32.6 & 4.68 & 38.3 & 66.0 & 14.0 & 95 \\
\rowcolor{mrOurs}
\textbf{NCE} & \textbf{31} & \textbf{34.8} & \textbf{3.70} & \textbf{30.5} & \textbf{61.1} & 3.1 & 99 \\
\bottomrule
\end{tabular}
\caption{\textbf{Five ways to manage context on Terminal-Bench 2.0 with LongCat under a 128K budget.} Solved counts are out of 89 tasks; tokens, context, reductions, and turns are per task.}
\label{tab:app_ctx_runs}
\end{table*}

\paragraph{Decoding.}
All methods sample with temperature 1.0, top-$p$ 0.95, and top-$k$ 50, with at most 32,768 output tokens per model call. Thinking is off in every run.

\section{Analysis}
\label{app:analysis}

This appendix details the ablations of Section~\ref{sec:ablation}: context management (\ref{app:ctx_mgmt}), building supervision (\ref{app:supervision}), the edits NCE writes at inference (\ref{app:inference}), and the cost of re-prefilling after an edit (\ref{app:prefill}).

\subsection{Context Management}
\label{app:ctx_mgmt}

\paragraph{Methods.}
All five methods run on Terminal-Bench 2.0 under a 128K budget. The four baselines use the LongCat base model from which NCE is trained.
\emph{Harness compaction} is the built-in mechanism of Terminus-2: once the history reaches the budget, the harness asks the model for a summary, lets a fresh instance ask questions about it, and continues from the summary, the questions, and their answers.
The other three baselines run inside the runtime in place of harness compaction and always keep the system prompt and the task description.
The \emph{sliding window} drops the oldest turns once the prompt exceeds the budget, down to 75\% of the budget, and leaves a one-line marker in their place.
The \emph{fixed memory state} drops turns the same way and adds a memory block of at most 4,096 tokens, which the base model rewrites from the previous memory and the dropped turns at every eviction.
\emph{Prompt compression} has the base model compress every tool output longer than 1,024 tokens to at most 256 tokens once it is older than the two most recent turns, and falls back to the sliding window when the prompt still exceeds the budget.
NCE keeps harness compaction as in Table~\ref{tab:tb_budgets}; since the harness counts its own copy of the history, it also compacts some NCE tasks.

\paragraph{Accounting.}
Tokens per task sum the prompt and output tokens of all model calls, including compaction calls. As in Table~\ref{tab:main_results}, tokens are averaged over tasks with a 10\% trimmed mean.
The context of a turn is the number of prompt tokens the model receives; we average the mean and the peak of each task over tasks.
A reduction is one harness handoff, one eviction of the sliding window or the memory state, one compression call, or one NCE turn with an executed edit.
Table~\ref{tab:app_ctx_runs} lists the results.

\subsection{Building Supervision}
\label{app:supervision}

\paragraph{Designs.}
We compare three ways to build editing supervision on 40 trajectories sampled from our long-trajectory training pool, stratified by source and by the number of editing decisions (26 terminal tasks and 14 software engineering issues, 97 decisions in total). All three use the same teacher model, decide at the same turns, and receive the same trigger signals; the decision turns are those at which the production pipeline edited. \emph{One teacher, one pass} reads the whole trajectory and writes the edits of every decision in a single response; its prompt states that the agent at decision turn $t$ has seen only the blocks and actions up to $t$ and requires every summary to use that information alone. \emph{Two teachers, original history} keeps the release points of the offline teacher and runs the online teacher at every decision on the unedited history. \emph{Two teachers, evolving context} is NCE: each decision is made on the context that the earlier edits produced. The two two-teacher designs share their first decision, since no edit precedes it, and differ only from the second decision on.

\paragraph{Defects.}
We execute the edits of each design in order on the recorded trajectory and check every edit against what the agent could see at that moment. A \emph{literal} is a path, file name, identifier with an underscore or an inner capital, number with at least four digits, decimal number, IPv4 address, or hexadecimal hash, of at least five characters, that occurs in fewer than 2\% of 1,000 reference trajectories; the frequency filter removes strings such as \texttt{os.path} that an agent can write from general knowledge. An edit at turn $t$
\begin{itemize}\itemsep0pt
\item \emph{uses future information} if its text contains a literal that first appears in the trajectory after turn $t$ and is absent from the action of turn $t$;
\item \emph{deletes evidence used later} if it removes from the context a literal that a later action uses before any later observation supplies it again;
\item \emph{edits a removed block} if it targets a block that is no longer in the context;
\item \emph{restates removed text} if its text contains a literal that occurs in the original history but no longer in the context.
\end{itemize}
A trajectory has a defect of a type if at least one of its edits does. Intervals are 95\% bootstrap intervals over trajectories, and $p$-values come from an exact paired sign test over trajectories.

\begin{table}[h]
\centering
\small
\setlength{\tabcolsep}{4pt}
\begin{tabular}{lccc}
\toprule
Trajectories (\%) & One pass & Original & NCE \\
\midrule
Uses future information & 30.0 & 12.5 & \textbf{5.0} \\
Deletes evidence used later & 72.5 & 40.0 & \textbf{37.5} \\
Edits a removed block & \textbf{0.0} & 95.0 & \textbf{0.0} \\
Restates removed text & 7.5 & 85.0 & \textbf{2.5} \\
Any defect & 80.0 & 100.0 & \textbf{37.5} \\
\midrule
Defective edits (\%) & 4.5 & 22.5 & \textbf{1.1} \\
Edits & 3,213 & 4,229 & 2,484 \\
\bottomrule
\end{tabular}
\caption{\textbf{Defects of three ways to build supervision on 40 trajectories.} Original: two teachers on the original history.}
\label{tab:supervision}
\end{table}

\paragraph{Results.}
Table~\ref{tab:supervision} lists all rates. Against the single pass, NCE has fewer trajectories with future information (2 against 12 discordant trajectories, $p=0.013$), with deleted evidence (2 against 16, $p=0.0013$), and with any defect (1 against 18, $p=7.6\times10^{-5}$). Against the original history, NCE has fewer trajectories with edits of removed blocks (0 against 38, $p=7\times10^{-12}$) and with restated removed text (0 against 33, $p=2\times10^{-10}$). The original-history design re-edits the blocks its earlier decisions already handled: 810 of its 4,229 edits target a removed block.

\paragraph{Drift.}
Figure~\ref{fig:supervision}b measures the drift on 4,250 annotated training trajectories. At each editing decision, the drift is the share of the tokens of the unedited history that are no longer in the agent's context, computed from the context lengths of the same trajectory replayed with and without its edits under the same renderer and tokenizer. Of the training trajectories, 75\% have at least two editing decisions, and 47\% of supervised turns follow the first edit. We exclude trajectories whose two replays already differ by more than 1\% before the first edit (2,640 of 6,890 sampled, 2,582 of them from a single data batch).

\subsection{Editing at Inference}
\label{app:inference}

\begin{figure*}[t]
  \centering
  \includegraphics[width=\textwidth]{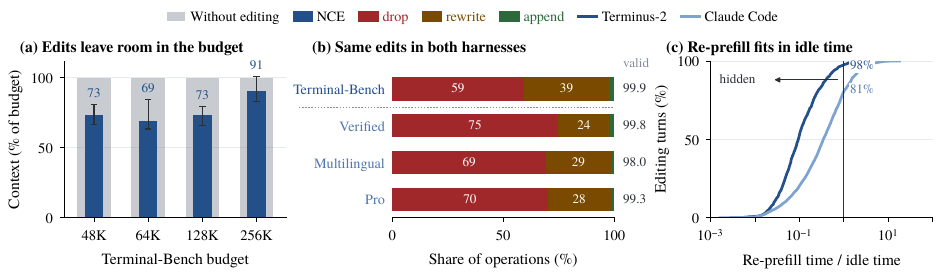}
  \caption{\textbf{Editing at inference.} (a)~Context NCE maintains when the unedited history first fills the budget on Terminal-Bench, as a share of the budget (median and interquartile range). (b)~Operation mix at 256K and share of edits executed as written. (c)~Cumulative share of editing turns by re-prefill time over idle time at 9.5K tokens/s; turns left of 1 hide the re-prefill entirely.}
  \label{fig:inference}
\end{figure*}

We read every turn the NCE runtime served in the LongCat runs of Tables~\ref{tab:main_results} and~\ref{tab:tb_budgets}. An operation is one envelope \tokctx\ldots\tokctxend; an editing turn contains at least one operation that executes. Table~\ref{tab:inference_behavior} reports each benchmark at 256K and Figure~\ref{fig:inference} summarizes the analysis.

\begin{table}[t]
\centering
\footnotesize
\setlength{\tabcolsep}{2.8pt}
\begin{tabular}{l cccc}
\toprule
 & TB2 & Verified & Multi. & Pro \\
\midrule
Tasks with edits                 & 89.7 & 93.2 & 97.0 & 93.4 \\
Edits without reminder           & 71.6 & 94.6 & 85.0 & 88.8 \\
Operations per task              & 32.0 & 36.1 & 55.8 & 48.0 \\
Editing turns per task           & 1.6  & 1.8  & 2.3  & 1.9  \\
Editing turns of all turns       & 2.9  & 2.0  & 1.7  & 1.5  \\
Editing turns of all output      & 3.8  & 9.0  & 8.7  & 8.9  \\
\opdrop{}                        & 59   & 75   & 69   & 70   \\
\oprewrite{}                     & 39   & 24   & 29   & 28   \\
\opappend{}                      & 2    & 2    & 2    & 2    \\
Removed per editing turn         & 12.7K & 7.9K & 11.1K & 16.7K \\
Executed as written              & 99.9 & 99.8 & 98.0 & 99.3 \\
Ops.\ losing evidence            & 0.15 & 1.21 & 1.23 & 1.35 \\
Peak context reduction           & 8.1  & 11.3 & 13.4 & 14.3 \\
\bottomrule
\end{tabular}
\caption{\textbf{Editing behavior of NCE at 256K, in percent except the two per-task counts and the tokens removed per editing turn.} On Terminal-Bench, peak context reduction covers the turns before the first harness compaction.}
\label{tab:inference_behavior}
\end{table}

\paragraph{Token cost.}
We count the tokens processed per task as the sum of prompt and output tokens over all model calls, including compaction calls, and average them over tasks with a 10\% trimmed mean. At 256K, NCE processes 64\% of the tokens of harness compaction on average across the four benchmarks, and on Terminal-Bench the ratio stays between 61\% and 74\% at every budget.

\paragraph{Editing behavior.}
Editing turns make up 1.7\% of all turns and 8.1\% of output tokens, and the edits themselves take 5.7\% of output tokens. A task receives 45 operations in 1.9 editing turns on average, \opdrop{} accounts for 71\% of operations, and an editing turn removes 12.7K tokens on average. Of the 73,148 edits the model writes, 99.1\% are well formed, address live blocks, and execute as written; the rest mostly address a block that an earlier edit already removed or repeat an edit verbatim. Tighter budgets raise the share of reminded edits: on Terminal-Bench, edits issued without a reminder fall from 72\% at 256K to 27\% at 48K, and editing turns rise from 2.9\% to 5.0\% of turns.

\paragraph{Compaction.}
Because the edited context grows more slowly, the harness rarely needs to compact. Under Claude Code at 256K, it compacts in 3 of the 1,531 SWE-bench tasks of NCE (0.2\%), against 14 of 1,531 under tool-call editing (0.9\%) at the same threshold.

\paragraph{Evidence.}
Edits rarely remove evidence the agent still needs. We trace the rare literals of every block an \opdrop{} or \oprewrite{} removes, with the literal definition of Appendix~\ref{app:supervision}, through the agent's later actions. Of the removed literals, 1.4\% reappear in a later action, and 0.45\% reappear while absent from the agent's context. Only 1.2\% of \opdrop{} and \oprewrite{} operations remove a literal that the agent later writes without seeing it, and 0.2\% when only paths, numbers, and hashes count. Most of these are framework names the model writes from knowledge, such as \texttt{django.core}, or paths a rewritten sub-agent report left out.

\paragraph{Budget.}
Edits free a quarter of the budget before compaction is due. On Terminal-Bench, Terminus-2 compacts once the unedited history reaches the budget, and NCE's context is always smaller than that history, so we compare the two only before the first harness compaction of each task, where neither side has been compacted. When the unedited history first fills a budget of 48K, 64K, or 128K, the context NCE maintains fills a median of 73\%, 69\%, and 73\% of it (Figure~\ref{fig:inference}a): by then NCE has freed over a quarter of the budget on its own. Over the same turns, edits lower the peak context of a task by a median of 19\%.

\paragraph{Harnesses.}
The edits keep their meaning across harnesses. Claude Code and Terminus-2 differ in prompts, tools, and observation formats, yet NCE edits in 94\% of SWE-bench tasks and 90\% of Terminal-Bench tasks, \opdrop{} leads the operation mix in both (71\% and 59\%), an editing turn removes 12.7K tokens in both, and 99.1\% and 99.9\% of edits execute as written (Figure~\ref{fig:inference}b). In Claude Code the removed blocks are tool results, led by shell output (31\%), file reads (22\%), and searches (16\%); the task itself is edited in under 1\% of operations.

\subsection{Re-prefill After an Edit}
\label{app:prefill}

An edit changes the rendered context from the first edited block onward, so the KV cache of everything after that block must be recomputed. Edits execute when the turn ends, and every token of the next prompt except the new observation is then known. A runtime can therefore issue the re-prefill as a separate request before the harness returns the observation. We measure how much time this re-prefill costs and how much of it fits into time the model server would otherwise leave idle for this agent.

\paragraph{Re-prefill length.}
For each editing turn we compare the next prompt with the sequence held in the prefix cache, the previous prompt followed by the model's output. The re-prefill spans from the first differing token to the start of the new observation; the observation itself is excluded because every method must prefill it. Turns without edits differ only in the turn delimiter, a single token, which confirms the comparison. The model edits old blocks, and the oldest blocks sit right after the task, so an edit invalidates most of the context: the median re-prefill is 23K to 58K tokens on Terminal-Bench (86--97\% of the prompt) and 13K to 22K tokens on SWE-bench (45--55\%, behind roughly 18K tokens of Claude Code tool declarations that are never edited).

\begin{table}[t]
\centering
\small
\begin{tabular}{l cc}
\toprule
Prefill rate & Terminus-2 & Claude Code \\
\midrule
5K tokens/s            & 92 & 62 \\
\textbf{9.5K tokens/s} & \textbf{98} & \textbf{81} \\
20K tokens/s           & 100 & 93 \\
40K tokens/s           & 100 & 98 \\
\midrule
Time hidden (9.5K)     & 98 & 81 \\
\bottomrule
\end{tabular}
\caption{\textbf{Share of editing turns (\%) whose re-prefill fits entirely in the idle time, at four prefill rates}; bold marks the rate measured from our traces. The last row gives the share of re-prefill time hidden.}
\label{tab:prefill}
\end{table}

\begin{figure}[t]
  \centering
  \includegraphics[width=\columnwidth]{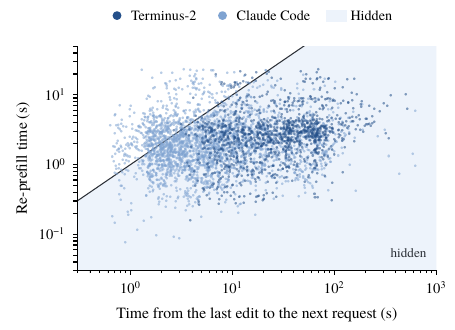}
  \caption{\textbf{Re-prefill time against idle time for every editing turn at 9.5K tokens/s.} Points in the shaded region are hidden entirely.}
  \label{fig:prefill_scatter}
\end{figure}

\paragraph{Prefill rate.}
The traces record when each request was sent and answered, but no time to first token. We regress the latency of every request on its prompt length, its output length, and its re-prefill length, with a separate intercept per run. Each re-prefilled token adds 0.105\,s per 1K tokens on Claude Code (95\% CI 0.083--0.123, bootstrap over tasks), an effective rate of 9.5K tokens/s under the load of our evaluation. An ordinary prompt token adds only 0.039\,s per 1K, which shows that ordinary turns reuse the prefix cache. On Terminus-2 the estimate is noisier (0.044\,s per 1K, CI $-0.001$--0.094), and we apply the Claude Code rate to both harnesses, which is conservative for Terminus-2.

\paragraph{Cost on our runs.}
Our runtime never overlaps the re-prefill with other work, so the measured cost is an upper bound. It adds 0.2--0.4\% to the wall-clock time of a task on every benchmark and budget (90th percentile at most 0.8\%), since a task has 1.6 to 5.4 editing turns among tens to hundreds of turns. In token terms, the re-prefill amounts to 0.7--4.8\% of the tokens processed per task, while the edits remove 4 to 22 times as many tokens from the prompts of later turns.

\paragraph{Hiding the re-prefill.}
A turn writes its edits before its action, so the runtime knows the new context as soon as the last edit closes, while the model still writes the action and the harness then executes it. The idle time available to the re-prefill therefore runs from the last edit of the turn to the next request. We estimate its decoding part from the output tokens after the last edit and the measured decoding time per token, and take the rest from the harness timestamps. Starting at the last edit, the re-prefill fits entirely in 98\% of editing turns under Terminus-2 and 81\% under Claude Code (Table~\ref{tab:prefill}, Figure~\ref{fig:prefill_scatter}). These shares are computed from the measured timings of our runs.

\clearpage
\onecolumn
\IfFileExists{Sections/appendix/E_case_study.tex}{\section{Case Studies}
\label{app:case}

Figure~\ref{fig:case_tb2} follows NCE through one Terminal-Bench 2.0 task at a 48K budget. NCE solves the task, and both baselines fail it. All excerpts are verbatim model output from the evaluation logs, shortened with \ldots.

\newcommand{\caseBadge}[3]{\tikz[baseline=(b.base)]\node[circle, draw=#2, fill=#3, text=#2, line width=0.5pt,
  inner sep=0pt, minimum size=1.15em, font=\scriptsize\bfseries](b){#1};}
\newcommand{\caseHandoff}{\tikz[baseline=-0.6ex]\draw[line width=0.6pt, dash pattern=on 0.6pt off 0.9pt] (0,-0.55em) -- (0,0.75em);}
\newcommand{\caseHead}[2]{\textbf{Turn #1}\enspace{\color{black!60}#2}\par}
\newcommand{\caseWhy}[1]{{\itshape ``#1''}\par}
\newenvironment{caseItem}[1]
  {\par\noindent\begin{minipage}[t]{1.5em}\vspace{0pt}#1\end{minipage}%
   \begin{minipage}[t]{\dimexpr\linewidth-1.5em\relax}\vspace{0pt}\raggedright}
  {\end{minipage}\par\vspace{4.5pt}}

\begin{figure*}[!ht]
\centering
\resizebox{\textwidth}{!}{\begin{minipage}{1.11\textwidth}
\begin{tcolorbox}[colback=black!3, colframe=black!35, boxrule=0.4pt, arc=1.5pt, left=5pt, right=5pt, top=3pt, bottom=3pt, fontupper=\small]
\textbf{Task} \texttt{tune-mjcf} (Terminal-Bench 2.0).\enspace
``Can you tune this MuJoCo model file (mjcf) such that it takes 60\% of the original time or less to simulate the same scene for a total of two simulation seconds? The same full physics state should be reached within atol=1e-5 without NaN or Inf. \ldots''\par\smallskip
\textbf{Harness} Terminus-2\enspace$\cdot$\enspace\textbf{Budget} $W=$ 48K tokens\enspace$\cdot$\enspace\textbf{Result}\enspace
Comp.~{\color{nceDropFg}\faTimes}\enspace Tools~{\color{nceDropFg}\faTimes}\enspace NCE~{\color{nceAppFg}\faCheck}
\end{tcolorbox}

\vspace{4pt}
\begin{minipage}[t]{0.405\textwidth}
  \vspace{0pt}
  \includegraphics[width=\linewidth]{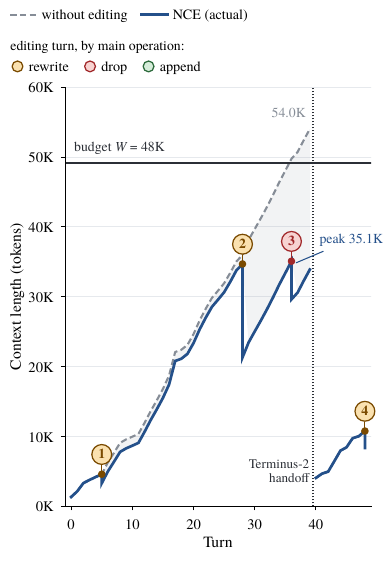}\par
  \vspace{6pt}\fontsize{8.5}{10.1}\selectfont
  \textbf{Totals}\par\vspace{2pt}\hrule\vspace{3pt}
  \renewcommand{\arraystretch}{1.12}
  \begin{tabular}{@{}l@{\hspace{6pt}}>{\raggedright\arraybackslash}p{\dimexpr\linewidth-4.45em\relax}@{}}
  Turns & 49 (4 with edits, 2 unprompted)\\
  Edits & 44, removing 22.6K tokens\\
        & \oprewrite{}\,30\enspace\opdrop{}\,11\enspace\opappend{}\,3\\
  Peak & 35.1K (71\% of $W$)\\
       & 54.0K (110\%) without editing\\
  Handoffs & 1 (Comp.\ and Tools: 3)\\
  Tokens & 0.90M processed (input and output)
  \end{tabular}
\end{minipage}\hfill
\begin{minipage}[t]{0.575\textwidth}
\vspace{0pt}\fontsize{8.5}{10.1}\selectfont
\textbf{Model output at the editing turns} {\color{black!60}(verbatim, shortened)}\par
\vspace{2pt}\hrule\vspace{4pt}

\begin{caseItem}{\caseBadge{1}{nceRewFg}{nceRewBg}}
\caseHead{5}{no reminder\enspace$\cdot$\enspace\oprewrite{}$\times$3\enspace$\cdot$\enspace$-$1.3K tokens}
\caseWhy{The setup and baseline phase is done, so I've condensed the initial file inspection, eval.py source, and baseline run into short notes that keep the key numbers, tolerances, \ldots}
\tokctx\oprewrite\tokaddr{00}{05} Baseline run with model.xml = exact copy of model\_ref.xml: final state difference 0.0000 (correctness passes), \ldots{} time pctg 0.9863 \ldots{} Conclusion: an unmodified copy is far too slow; the plan is to increase the simulation timestep \ldots\tokctxend
\end{caseItem}

\begin{caseItem}{\caseBadge{2}{nceRewFg}{nceRewBg}}
\caseHead{28}{soft reminder\enspace$\cdot$\enspace\opappend{}$\times$1, \oprewrite{}$\times$20\enspace$\cdot$\enspace$-$13.4K tokens}
\caseWhy{The tuning sweep and error-formatting experiments have served their purpose, so I've condensed them into a compact status note that preserves the current model state, the invalid XML keywords, and the final-state accuracy measurements.}
\tokctx\opappend{} Consolidated tuning status \ldots: final-state differences observed vs model\_ref.xml, all exceeding atol=1e-5: timestep=0.01 Newton 500 iters tolerance=1e-6 $\to$ 0.6103; timestep=0.01 integrator=RK4 $\to$ unstable (qvel/qacc NaN/Inf at t=0.01, diff 6.47e9); \ldots{} current active variant: timestep=0.005 integrator=implicit $\to$ 0.00445. \ldots\tokctxend\par
\tokctx\oprewrite\tokaddr{00}{18} Explored mujoco module internals (MjSolverStat class, mjNSOLVER=200, mjtSolver enum); no further constants or API changes needed.\tokctxend\enspace{\color{black!60}(was a 3.4K-token \texttt{dir(mujoco)} listing)}
\end{caseItem}

\begin{caseItem}{\caseBadge{3}{nceDropFg}{nceDropBg}}
\caseHead{36}{soft reminder\enspace$\cdot$\enspace\opappend{}$\times$2, \opdrop{}$\times$11\enspace$\cdot$\enspace$-$5.5K tokens}
\caseWhy{The exploration phase has left a pile of superseded model experiments, so I've consolidated them into one status note \ldots, clearing out the dead-end logs while keeping every number and setting that might still matter.}
\tokctx\opappend{} Consolidated solver/timestep tuning log \ldots: CG, timestep=0.002, jacobian=sparse, iterations=50, ls\_iterations=20 $\to$ correctness 0.0000, avg sim 0.3069 vs ref 0.3821, time pctg 80.31\%; \ldots{} No variant has yet reached time pctg $\le$ 0.6 while passing correctness; best candidates pass correctness at pctg $\sim$80\%.\tokctxend\par
\mbox{\tokctx\opdrop\tokaddr{00}{19}\tokctxend} \ldots{} \mbox{\tokctx\opdrop\tokaddr{01}{05}\tokctxend}\par
{\color{black!60}(11 drops: raw output of seven \texttt{model.xml} trials and four API probes)}
\end{caseItem}

\begin{caseItem}{\hspace{0.45em}\caseHandoff}
\caseHead{40}{Terminus-2 handoff}
Terminus-2 counts its own copy of the history, which still holds the 20.1K tokens NCE removed. That copy reaches the budget, so the harness summarizes and restarts the conversation; block addresses restart with it.
\end{caseItem}

\begin{caseItem}{\caseBadge{4}{nceRewFg}{nceRewBg}}
\caseHead{48}{no reminder\enspace$\cdot$\enspace\oprewrite{}$\times$7\enspace$\cdot$\enspace$-$2.4K tokens}
\caseWhy{The optimization task is complete (the PGS solver configuration passed eval.py twice with 50-52\% of reference time), so I have compressed the intermediate exploration logs and failed solver attempts into brief conclusions \ldots}
\tokctx\oprewrite\tokaddr{00}{09} Final validated configuration, written to /app/model.xml: \ldots{} solver=PGS, iterations=500, tolerance=1e-6, ls\_iterations=10, integrator=Euler \ldots{} speedup 2.01x, time pctg 0.50 --- already passed both correctness and speed.\tokctxend
\end{caseItem}
\end{minipage}

\vspace{2pt}
\end{minipage}}
\vspace{2pt}
\caption{\textbf{NCE keeps a Terminal-Bench task inside a 48K budget by folding finished experiments into notes.} Left: context length per turn. The dashed line adds back the tokens removed by earlier edits; the shaded gap is what NCE keeps out of the context. Numbered markers are editing turns, colored by the operation that removes the most tokens. Right: rationale (italics) and edits at each editing turn, with the net tokens the turn removes. Each failed configuration becomes one line with its settings and error, and the raw logs go once a note covers them. Terminus-2 counts the unedited history and hands off at turn~40 (dotted line).}
\label{fig:case_tb2}
\end{figure*}
}{}
\IfFileExists{Sections/appendix/E2_case_swe.tex}{Figure~\ref{fig:case_swe} shows the same model in Claude Code on SWE-bench Verified at a 256K budget, where it edits without any reminder.

\providecommand{\caseBadge}[3]{\tikz[baseline=(b.base)]\node[circle, draw=#2, fill=#3, text=#2, line width=0.5pt,
  inner sep=0pt, minimum size=1.15em, font=\scriptsize\bfseries](b){#1};}
\providecommand{\caseHead}[2]{\textbf{Turn #1}\enspace{\color{black!60}#2}\par}
\providecommand{\caseWhy}[1]{{\itshape ``#1''}\par}
\ifcsname caseItem\endcsname\else
\newenvironment{caseItem}[1]
  {\par\noindent\begin{minipage}[t]{1.5em}\vspace{0pt}#1\end{minipage}%
   \begin{minipage}[t]{\dimexpr\linewidth-1.5em\relax}\vspace{0pt}\raggedright}
  {\end{minipage}\par\vspace{4.5pt}}
\fi

\begin{figure*}[!ht]
\centering
\begin{tcolorbox}[colback=black!3, colframe=black!35, boxrule=0.4pt, arc=1.5pt, left=5pt, right=5pt, top=3pt, bottom=3pt, fontupper=\small]
\textbf{Task} \texttt{django\_\_django-14855} (SWE-bench Verified).\enspace
``Wrong URL generated by get\_admin\_url for readonly field in custom Admin Site \ldots''\par\smallskip
\textbf{Harness} Claude Code\enspace$\cdot$\enspace\textbf{Budget} $W=$ 256K tokens\enspace$\cdot$\enspace\textbf{Result} NCE~{\color{nceAppFg}\faCheck}
\end{tcolorbox}

\vspace{4pt}
\begin{minipage}[t]{0.405\textwidth}
  \vspace{0pt}
  \includegraphics[width=\linewidth]{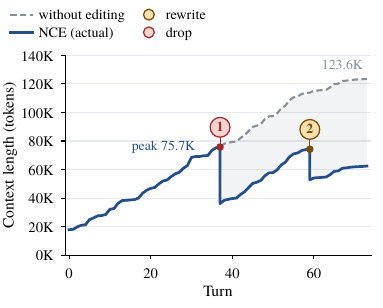}\par
  \vspace{6pt}\fontsize{8.5}{10.1}\selectfont
  \textbf{Totals} {\color{black!60}(main agent)}\par\vspace{2pt}\hrule\vspace{3pt}
  \renewcommand{\arraystretch}{1.12}
  \begin{tabular}{@{}l@{\hspace{6pt}}>{\raggedright\arraybackslash}p{\dimexpr\linewidth-4.45em\relax}@{}}
  Turns & 74 (2 with edits, both unprompted)\\
  Edits & 54, removing 61.1K tokens\\
        & \oprewrite{}\,23\enspace\opdrop{}\,30\enspace\opappend{}\,1\\
  Peak & 75.7K; 123.6K without editing
  \end{tabular}
\end{minipage}\hfill
\begin{minipage}[t]{0.575\textwidth}
\vspace{0pt}\fontsize{8.5}{10.1}\selectfont
\textbf{Model output at the editing turns} {\color{black!60}(verbatim, shortened)}\par
\vspace{2pt}\hrule\vspace{4pt}

\begin{caseItem}{\caseBadge{1}{nceDropFg}{nceDropBg}}
\caseHead{37}{no reminder\enspace$\cdot$\enspace\opappend{}$\times$1, \oprewrite{}$\times$3, \opdrop{}$\times$30\enspace$\cdot$\enspace$-$39.6K tokens}
\caseWhy{The reproduction phase just converged, so I've tidied up in one pass: the earlier iterations of the reproduction script, their errors and dead ends are now condensed into a single summary note (keeping the key paths, commands and the root-cause finding), while the current script, the source analysis and the test fixtures stay untouched.}
\tokctx\opappend{} Reproduction-phase notes (consolidated): \ldots{} Dead ends hit while building it: importing django.contrib.auth.models.User with no configured settings raises RuntimeError (app\_label); \ldots{} The final converged run confirmed the root cause: with no current\_app, reverse() fails to resolve the custom admin site's URL, so AdminReadonlyField.get\_admin\_url falls through the NoReverseMatch handler \ldots\tokctxend\par
\mbox{\tokctx\opdrop\tokaddr{00}{03}\tokctxend} \ldots{} \mbox{\tokctx\opdrop\tokaddr{01}{06}\tokctxend}\par
{\color{black!60}(30 drops, mostly earlier versions of the reproduction script and their failing test runs)}
\end{caseItem}

\begin{caseItem}{\caseBadge{2}{nceRewFg}{nceRewBg}}
\caseHead{59}{no reminder\enspace$\cdot$\enspace\oprewrite{}$\times$20\enspace$\cdot$\enspace$-$21.5K tokens}
\caseWhy{Let me condense the intermediate material I'm done with, keeping the key findings, then carry on.}
\tokctx\oprewrite\tokaddr{01}{10} First edit to /testbed/django/contrib/admin/helpers.py get\_admin\_url: added current\_app=self.admin\_site.name to the reverse() call (around line 215). Superseded --- the current code uses current\_app=self.model\_admin.admin\_site.name instead.\tokctxend\par
\tokctx\oprewrite\tokaddr{01}{12} Final implemented fix in AdminReadonlyField.get\_admin\_url \ldots{} This is the change that passes the reproduction script and that the upcoming regression/edge-case verification must cover.\tokctxend
\end{caseItem}
\end{minipage}

\caption{\textbf{The same model edits its context the same way under Claude Code.} Both editing turns come without a reminder. The first folds a finished reproduction phase into one note with the dead ends and the root cause, then drops the superseded scripts and failing runs; the second rewrites spent file reads and edits into one-line records, keeping which fix was replaced and which one is final. Plot and excerpt conventions follow Figure~\ref{fig:case_tb2}.}
\label{fig:case_swe}
\end{figure*}
}{}

\end{document}